\pdfoutput=1

\documentclass[11pt]{article}

\usepackage{acl}  
\usepackage{multicol}
\usepackage{listings}
\usepackage{xcolor}
\definecolor{deepgreen}{RGB}{50,220,50}

\usepackage{times}
\usepackage{latexsym}
\usepackage{todonotes}
\usepackage{multirow}
\usepackage{subcaption}
\usepackage{pifont}
\usepackage{amssymb}
\usepackage{amsmath}
\usepackage[most]{tcolorbox}
\usepackage[T1]{fontenc}

\usepackage[utf8]{inputenc}

\usepackage{microtype}

\usepackage{inconsolata}

\usepackage{booktabs}
\usepackage{array}
\usepackage{caption}
\usepackage[normalem]{ulem}

\title{ProST: Progressive Sub-task Training for Pareto-Optimal Multi-agent Systems Using Small Language Models}

\author{
\textbf{Biddut Sarker Bijoy}\textsuperscript{1} \quad
\textbf{Mohammad Saqib Hasan}\textsuperscript{1} \quad
\textbf{Pegah Alipoormolabashi}\textsuperscript{1} \\
\textbf{Avirup Sil}\textsuperscript{2} \thanks{Work done while at IBM Research AI.}\quad
\textbf{Aruna Balasubramanian}\textsuperscript{1} \quad
\textbf{Niranjan Balasubramanian}\textsuperscript{1} \\
\textsuperscript{1}Stony Brook University \quad
\textsuperscript{2}Oracle Applied Science \\
\textsuperscript{1}\{bbijoy, mdshasan, palipoormola, arunab, niranjan\}@cs.stonybrook.edu, \;
\textsuperscript{2}avi.sil@oracle.com}

\begin{document}
\newcommand{\pegah}[2][]{\textcolor{blue}{\texttt{[Pegah]} #2}}
\newcommand{\bijoy}[2]{\textcolor{deepgreen}{\texttt{[Bijoy]} #2}}
\newcommand{\arr}[1]{\textcolor{purple}{#1}}
\newcommand{\saqib}[2]{\textcolor{brown}{\texttt{[Saqib]} #2}}
\newcommand{\nb}[2][]{\textcolor{red}{\texttt{[Niranjan]} #2}}
\newcommand{\avi}[2][]{\textcolor{Olive}{\texttt{[Avi]} #2}}
\newcommand{\aruna}[2][]{\textcolor{magenta}{\texttt{[Aruna]} #2}}

\newcommand{\eat}[1]{}

\newcommand{\orchestrator}[0]{Orchestrator}
\newcommand{\critic}[0]{Critic}
\newcommand{\exec}[0]{Executor}

\newcommand{\cmark}{\ding{51}}
\newcommand{\xmark}{\ding{55}}

\newcommand{\Algo}{Progressive Subtask Training}
\newcommand{\algo}{\texttt{ProST}}
\newcommand{\appworld}{\texttt{AppWorld}}
\newcommand{\loss}[0]{\algo}

\newcommand{\base}[0]{\texttt{SA}}
\newcommand{\saft}[0]{\texttt{[FT]}}
\newcommand{\mas}[0]{\texttt{MA}}
\newcommand{\masft}[0]{\texttt{[FT]}}
\newcommand{\masalgo}[0]{\texttt{[\algo{}]}}

\newcommand{\tgc}[0]{\texttt{TGC}}
\newcommand{\sgc}[0]{\texttt{SGC}}
\newcommand{\flops}[0]{\texttt{FLOPs}}
\newcommand{\react}[0]{\texttt{ReAct}}

\maketitle
\begin{abstract}

Multi-agent systems with smaller language models (SLMs) present a viable alternative to single agent systems powered by large language models (LLMs) for addressing complex problems. In this work, we study how these alternatives compare in terms of both effectiveness and efficiency. To study this trade-off, we instantiate single and multi-agent systems for the complex problems in the AppWorld environment using different sized language models.

We find that difficulties with long-trajectory learning in smaller language models (SLMs) limit their performance. Even when trained for specialized roles, SLMs fail to learn all subtasks effectively. To address this issue, we introduce a simple progressive sub-task training strategy, which introduces new sub-tasks progressively in each training epoch. We find that this novel strategy, analogous to instance level curriculum learning, consistently improves the effectiveness of multi-agents at all configurations. Our Pareto analysis shows that fine-tuned multi-agent systems yield better effectiveness-efficiency trade-offs. Additional ablations and analyses shows the importance of our progressive training strategy and its ability to reduce subtask error rates.


\end{abstract}

\section{Introduction}
Solving complex problems requires a wide-range of capabilities including planning sub-tasks, acting in an environment via coding, reasoning about the actions, recovering from errors, as well as tracking and managing the execution of sub-tasks. LLMs with their strong coding and reasoning abilities and advances in handling long contexts have shown impressive breakthroughs on such complex problems (e.g., \appworld{}~\citep{appworld-acl24} and \texttt{SWE-Bench}~\citep{jimenez2024swebench}). However, these large models incur high computational and API costs for both training and inference. 
Another viable alternative is to use multi-agent solutions with smaller language models\footnote{Small and large are subjective terms. For this paper we use SLM for models with fewer than 40B parameters.} (SLMs). The core idea is to identify specialized roles in solving these complex tasks (e.g., orchestration, code writing, and critiquing) and use separate SLMs (or SLM calls) for each role. \citet{belcak2025smalllanguagemodelsfuture} argue that SLMs are much more cost-effective and practical for training, adapting, and deploying multiple specialized experts for different agentic routines. ~\citet{shen2024small} show these multi-agent solutions with SLMs can be effective for complex problems requiring tool use via APIs.

We consider two challenges that arise in this context. First, for SLMs, even specialized roles can be difficult to learn. Complex problems often have multiple subtasks resulting in long trajectories. Despite their improving capabilities, SLMs still struggle to learn all subtasks under standard fine-tuning. Second, while the per-token compute depends on the size of the models, the overall computational cost also depends upon the nature of their solutions (e.g., their length). Therefore, it is important to explicitly consider both the effectiveness and efficiency of these solutions to better understand the trade-offs. 

We address these challenges using \appworld{} benchmark. We design a multi-agent solution with three agents: an \orchestrator{} that plans subtasks incrementally, an \exec{} that interacts with the environment via code to solve subtasks, and a \critic{} agent that critiques the \exec{} output. We make two specific contributions on this multi-agent setup:

\textbf{1) Progressive Sub-task Training} To address the inconsistent performance on large trajectories, we introduce a progressive training strategy. We find that SLMs are unable to solve some sub-tasks of these complex problems, even in role-specialized learning settings. We propose a simple curriculum-style learning strategy, where we introduce sub-tasks progressively in training epochs. This allows the small capacity SLMs to gradually expand their learning, covering all aspects of the problem trajectories.


\textbf{2) Pareto Analyses} We compare different instantiations of single and multi-agent systems. In particular, we plot the efficacy of different solutions against inference-time \flops{}. The Pareto curve for these solutions helps choose the best solution at different compute budgets. 

We evaluate these ideas using \texttt{Qwen-2.5 Coder}, \texttt{Llama-3.1} and \texttt{Phi-4} models. 
We distill training trajectories for a single agent from a frontier model (\texttt{Claude-3.7-Sonnet}). Then, we translate these into role-specific multi-agent training trajectories for three agents, namely, an \orchestrator{} (planner), an \exec{} (coding) agent, and a \critic{} agent. Our evaluations yield multiple interesting observations \appworld{} tasks: (i) While larger models tend to have higher effectiveness, multi-agent solutions can achieve better efficiency versus effectiveness trade-offs. (ii) \Algo{} yields a better Pareto-front compared to standard fine-tuning. (iii) Stronger (higher compute) agents for planning are more effective (and efficient) than using stronger coding and critic agents when finetuning using \Algo{}.\footnote{Code and data available at https://github.com/StonyBrookNLP/prost-multiagents}





\section{Related Work}




\paragraph{SLM-based Agents}
\vspace*{-2mm}
SLMs are being used as agents due to lower computational cost, but struggle with complex planning and tool use \citep{shen2024small}, skills essential for agents. Hence, many strategies are proposed: separation of roles to only tackle small, concrete tasks \citep{shi-etal-2024-learning,qiao-etal-2024-autoact}; distilling training data from powerful LLMs\citep{shi-etal-2024-learning}, Direct Policy Optimization (DPO)\citep{feng2025empoweringllmstaskorienteddialogues}; online learning by generating data from unsuccessful attempts\citep{qi2025webrltrainingllmweb}, masking loss from erroneous steps\citep{fu2025agentrefineenhancingagentgeneralization}; learning only from critical steps \citep{chen2025atlasagenttuninglearning}. 
Contrary to these works, we formulate \algo{} which optimizes SLMs, as part of multi-agent design, by initiating with important parts of the trajectory and iteratively increasing steps, a strategy never tried for agents.



\paragraph{Multi-Agent Optimizations} Many works propose techniques to make multi-agents more robust. These techniques include dynamically reducing redundant agents and messages \citep{wang2025agentdropoutdynamicagentelimination,zhang2024cutcrapeconomicalcommunication}, iterative answer refinement through feedback\citep{chen2025magicore}, optimizing collaboration strategies\citep{wang-etal-2025-beyond, zeng-etal-2025-s2} and communication protocols \citep{xiao2025longcontextscalingdivide}, and test-time search for effective tool planning \citep{chen-etal-2025-smurfs}. Recent works propose using cost-effective SLMs for multi-agent setups: \citet{erdogan2025planandactimprovingplanningagents} uses smaller\texttt{32-B} models as planning and execution subagents; \citet{belcak2025smalllanguagemodelsfuture} argue for SLM multi-agents based on their improved capabilities, cost, ease of adaptability, and deployment. 
Our work introduces \algo{}, a training algorithm for SLMs in multi-agent setups that iteratively trains on subtasks. Unlike previous works, we show how \algo{} produce SLMs effective at solving tasks as well as computationally efficient.

\paragraph{Efficiency of Multi-agents}
Recent works have sought to assess and improve the computational efficiency of multi-agent systems. 
For example, \citet{wang2025agentdropoutdynamicagentelimination} use token consumption as a metric to prune redundancies in multi-agent collaborations. \citet{yue-etal-2025-masrouter} show, via Pareto fronts, that their language model routing strategy in multi-agent systems reduces dollar cost while improving accuracy. \citet{wang-etal-2025-beyond} use \textit{token accuracy ratio} to compare multi-agent collaboration strategies.
In contrast, we analyze computational cost against task accuracy to measure \textit{Pareto optimality}. Based on this, we design \algo{}, which gives better performance accuracy with higher computational efficiency (more \textit{Pareto optimal}).

\section{Multi-Agent Design for AppWorld}
\label{sec:methodology}
Multi-agent systems using SLMs instead of LLMs have emerged as an alternative for better cost-performance trade-off~\cite{erdogan2025planandactimprovingplanningagents,belcak2025smalllanguagemodelsfuture}. However, optimizing SLMs for long and complex reasoning tasks remains challenging\citep{shen2024small}. In this work, we focus on \appworld{} ~\citep{appworld-acl24}, 
an agent environment of complex tasks requiring coding and ability to interact with APIs. Our method is based on two considerations: (i) SLMs are less effective at essential agent skills like coding, long-context and recovering from errors; (ii) overall computation cost is based on both solution length  and low per-token cost of SLMs.



To address these challenges we: (i) use a tri-agent system for \appworld{} (\autoref{sec:architecture}), (ii) introduce \Algo{}, a novel training loss that improves SLMs on complex long trajectory reasoning ~(\autoref{sec:pst}), and (iii) use Pareto analysis to understand the cost-performance trade-offs in agent systems~(\autoref{sec:experimental_setup}).

\begin{figure}
    \centering
    \includegraphics[width=\linewidth]{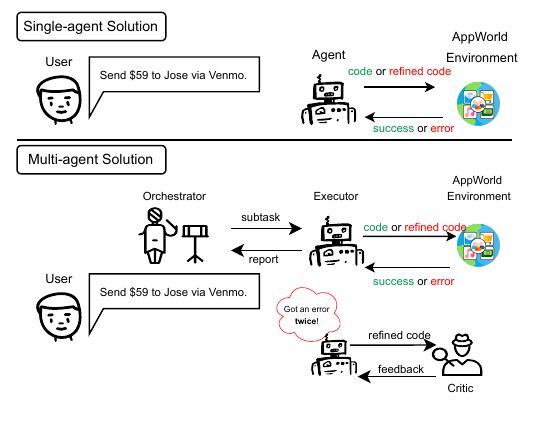}
    \caption{
        Comparison of single-agent and multi-agent architectures for solving complex tasks. The single-agent approach (top) employs a monolithic LLM that directly processes user given task and interacts with the \appworld{} environment by generating code and processing the output. The multi-agent approach (bottom) decomposes this functionality into three specialized components: an \orchestrator{} decomposes the user query into subtasks (one subtask at a time), an \exec{} that solve the given subtask step by step (one step at a time) by interacting with environment, and a \critic{} that evaluates and provides natural language feedback on the Executor's outputs when required. 
    }
    \label{fig:enter-label}
\vspace{-1.5em}
\end{figure}

\subsection{Multi-agent Architecture}
\label{sec:architecture}
Solving \appworld{} tasks require identifying sub-tasks that can solve each complex problem via writing code and self-reflection \cite{madaan2023self}.
Agents in multi-agent systems work in collaboration to solve such tasks, where each has its specific responsibilities. Such architectures have been used in \citet{qiao-etal-2024-autoact} (plan, tool, and reflect agents) and \citet{shi-etal-2024-learning} (grounding, execution, and review agents).
Accordingly, our multi-agent system (refer to Figure~\ref{fig:enter-label}) also comprises of similar elements: an \orchestrator{} that generates and delegates sub-tasks, an \exec{} that writes code to interact with \appworld{}, and a \critic{} that provides feedback to \exec{} (see prompt in Figure \ref{prompt:orchestrator}).

\paragraph{\orchestrator{}} This agent decomposes the task into subtasks with specific goals and execution plan. \orchestrator{} operates iteratively, executing one subtask after another and handling execution failures by \textit{dynamic decomposition strategy},i.e., further decomposing the subtask \citep{prasad2024adaptasneededdecompositionplanning}.

\paragraph{\exec{}} Agent responsible for solving each subtask generated by \orchestrator{}. It does so by writing code according to the plan, executing in \appworld{} and observing the results (see prompt in Figure \ref{prompt:executor}). Each subtask may need multiple iterations of such cycle. In case of error, \exec{} refines code. If error still persists it revises the code with feedback from \critic{}. Execution stops when either task is complete or maximum number of steps is reached.

\paragraph{\critic{}} Agent that acts when \exec{} fails during refinement. When prompted by \exec{} (see Figure \ref{prompt:critic}), \critic{} reviews the main task, subtasks, and the full trajectory of the \exec{} so far, as well as the \exec{}'s refinement. \critic{} then provides natural language feedback on the \exec{}'s suggested resolution.

\section{\Algo}
\label{sec:pst}
\vspace*{-4mm}
We fine-tune SLMs using multi-agent training trajectories derived from LLMs (described in \autoref{subsec:training-trajectories}). Standard fine-tuning trains each agent to maximize likelihood of the role relevant gold trajectory. However, as we metioned, the trajectories remain long, even with role specialization. SLMs' inability to process long trajectories~\cite{chen2025atlasagenttuninglearning} limits training them for their respective roles. In particular, we observe that standard finetuning leads to higher error rates for the more difficult middle subtasks compared to first and last subtasks. This observation motivates our proposed solution, drawing from curriculum learning.

\begin{figure*}
    \centering
\includegraphics[width=0.8\linewidth]{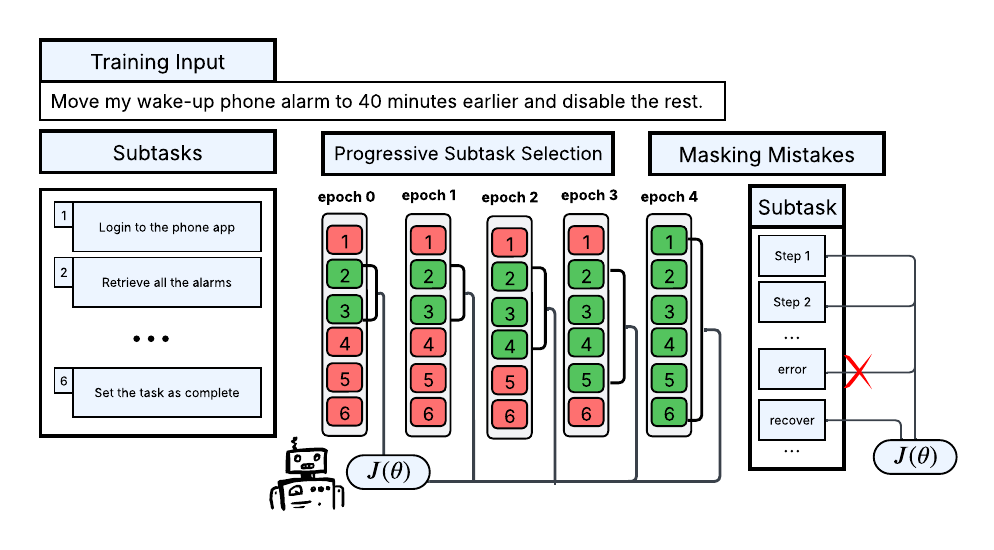}
    \caption{The figure shows how \algo{} trains models by progressively increasing the subtasks (highlighted in green) in training trajectory. The agent is trained on subtasks $2$ and $3$ in epoch $0$, ignoring subtask $1$ as it is not important(login tasks in \appworld{}). With each increasing epoch, subtasks presented to the model for training is increased from subtask $2$ onwards. In the final epoch, the full trajectory with all the subtasks are presented. Any step of a subtask that is erroneous is ommitted during training. In this way, model learns to: i) to solve the most important subtasks first, ii) sequentially learn to solve the larger problem at hand, and iii) avoid optimizing for solving subtasks erroneously.   
}
    \label{fig:prost}
\vspace{-1em}
\end{figure*}

Our approach is to use incremental curriculum on subtasks of each training instance. This ensures models train on subtasks subset in each iteration. We achieve this by randomly focusing on a set of subtasks at each epoch. However, later subtasks often depend on outputs of earlier ones, thereby requiring either understanding those tasks or accessing their output. For example, in task to send text message, selecting contacts presupposes retrieval of contact list. This dependency suggests that introducing subtasks into the curriculum in their natural order within the tasks is more effective, where the models learn the earlier subtasks first, followed by the later ones. 
\begin{align}
     &J^{(e)}_{O}(\theta, x) \nonumber \\
     &= \sum_{i=1}^{M_x} \mathbb{I}(i \in S^{(e)}) \ log(\pi_{\theta}(s_i|I,\left\{s_j, r_j \right\}_{j < i}))\\
     &J^{(e)}_{O}(\theta) = \mathbb{E}_{x \sim D_{O}} \left[ J^{(e)}_{O}(\theta, x) \right]
     \label{eq:orch}
\end{align}

This increment, however, may not be equal at each iteration as number of subtasks may not be perfectly distributable across the iterations. Hence, our algorithm tries to optimize subtask increment at later epochs. This is highlighted in Figure \ref{fig:prost} where first two epochs have no increment in subtasks but epochs $2$ to $4$ do, as $6$ subtasks cannot be divide into $5$ epochs for equal increment across epochs. (Further details about how subtasks are expanded progressively in Appendix \ref{appendix:prost_details}). Below, we formally describe this progressive subtask learning strategy.

\paragraph{\orchestrator{} Training} The \react{}-style training instances for the \orchestrator{} ($x \sim D_{Orch}$) include either (1) task descriptions as inputs paired with their corresponding initial subtasks as outputs, or (2) the \exec{} agent’s latest response as input, with the resulting subtasks and associated plans as the \texttt{action} outputs.
Suppose the entire trajectory for the \orchestrator{} consists of $m$ steps (i.e., \react{} turns) for task $x$, denoted by 
$(\{\phi\}, s_1),$ $(\{(s_1, r_1)\}, s_2),$ $\cdots,$ $(\{(s_1, r_1),$ $\cdots, (s_{m-1}, r_{m-1})\}, s_{m})$. For each turn $i$, $s_i$ denotes the subtask generated at step $i$, and the input to this step is the history of subtasks along with \exec{}'s final reports $\{(s_j, r_j)\}_{j < i}$ from previous steps. Here, $r_j$ represents the final report about the success/failure of the subtask $s_j$. 
Given $S^{(e)}$ the set of subtasks that the model is allowed to train on at epoch $e$ under the progressive training strategy, we can formally state the training objective for the \orchestrator{} (see eq \ref{eq:orch}). While standard fine-tuning trains over the entire trajectory at all epochs, our approach is to only train on progressively increasing parts of this trajectory.

\textbf{\exec{} and \critic{} Training} The \exec{} agent is trained to take as input a subtask and the corresponding plan $s_i$ generated by the \orchestrator{} and is expected to generate the sequence of turns of code. Additionally, the trajectory for the \exec{} agent includes environment feedbacks and interactions with the \critic{}. Suppose the entire trajectory for the overall task $x$ consists of $n$ steps (i.e. \react{} steps) and is given by 
$(\{\phi\}, (T_1, A_1, O_1)),$ $(\{(T_1, A_1, O_1)\}, (T_2, A_2, O_2)),$ $\cdots,$ $(\{(T_1, A_1, O_1),$ $\cdots,$ $(T_{m-1}, A_{m-1}, O_{m-1})\},$ $(T_{m}, A_{m}, O_{m}))$. For a given subtask $s_i$ at each step $t$, the input is the history of the conversation so far: $\{(T_j, A_j, O_j)\}_{j < t}$. Here $T$, $A$, and $O$ denote thoughts, actions, and observations.




In this learning paradigm, we train agents using both correct steps and self-corrected steps. This helps develop a comprehensive understanding of both successful subtask completion and error correction patterns. 
Following~\cite{fu2025agentrefineenhancingagentgeneralization} we compute the loss function $J^{(e)}_{E}$ (eq. \ref{eq:exec}), $J^{(e)}_{C}$ (eq. \ref{eq:critic}) only on correct and self-refined steps for \exec{} and \critic{}, respectively. In equation~\ref{eq:exe_cri}, $N_x$ is the total number of turns in a given task $x$. 
\begin{align}
     &J^{(e)}_{EC}(\theta, x) = \sum_{t=1}^{N_x} {\mathbb{I}(A_t).\mathbb{I}(s_t \in S^{(e)})}.\nonumber\\&log(\pi_{\theta}(T_t,A_t|I, subtask,\left\{  {T_j,A_j,O_j}\right\}_{j<t}))
     \label{eq:exe_cri}\\
     &J^{(e)}_{E}(\theta) = \mathbb{E}_{x \sim D_{E}} \left[ J^{(e)}_{EC}(\theta, x) \right]
     \label{eq:exec}\\
     &J^{(e)}_{C}(\theta) = \mathbb{E}_{x \sim D_{C}} \left[ J^{(e)}_{EC}(\theta, x) \right]
     \label{eq:critic}
\end{align}
\section{Experimental Setup}
\label{sec:experimental_setup}
We evaluate the performance of \Algo{} (\S\ref{s:pareto}), on \appworld{}, a popular agentic benchmark. We first \textit{data synthesis} to generate training trajectories for \appworld{} and then train and evaluate models and baselines using \appworld{} established metrics.
\subsection{AppWorld}
\appworld{} offers a testbed of day-to-day digital tasks that test agent abilities to solve complex problems using interactive coding, often over long trajectories. \appworld{}'s \textbf{execution engine} simulates 9 day-to-day apps: Amazon, Spotify, Venmo, Gmail, Todoist, SimpleNote, Splitwise, Phone, and FileSystem. \appworld{}'s \textbf{benchmark} is a dataset of $750$ realistic task instructions across $250$ different scenarios/use-cases, divided into $105$ train, $60$ dev, $105$ in distribution test and $417$ out-of-distribution test sets\footnote{We found only 90 train and 57 dev tasks are available in the appworld when we load the datasets.}. \appworld{} uses \textbf{state-based programmatic evaluation} by checking engine's final database state with manually-written assertions (details in~\autoref{appendix:appworld_details}).

\subsection{Data Synthesis}
\label{subsec:training-trajectories}
\appworld{} only provides training pairs (input prompts, output state). Hence, to obtain training trajectories, we design a \textit{data synthesis} pipeline similar to~\citep{erdogan2025planandactimprovingplanningagents} that first generates \react{}~\cite{react}-style trajectories for single agents and then uses a powerful LLM to translate them to multi-agent trajectories. First, we use 
\texttt{Llama-3.3-70B-Instruct}~\cite{llama2025instruct33} to generate single-agent trajectories (see the prompt in Figure~\ref{prompt:single_agent_traj_generation} in Appendix~\ref{appendix:prompts}) from \appworld{} train and dev sets via rejection sampling at different temperatures~\cite{chen2025reinforcementlearninglonghorizoninteractive}, obtaining  $2,844$ gold trajectories. Then, for each agent (\orchestrator{},\exec{} and \critic{}) in our multi-agent setup, we prompt \texttt{Claude 3.7-Sonnet}~\cite{anthropic2025claude37} to convert the single agent trajectories into appropriate trajectories for each agent (See the prompt in Figure~\ref{fig:sat_to_mat} in Appendix~\ref{appendix:prompts}). These trajectories were then validated using \appworld{}. We gathered $2,708$ such trajectories for training (see Appendix \ref{appendix:single-agent-data-creation} and \ref{appendix:sat-to-mat} for details).

\subsection{Models \& Setup}
\label{subsec:model_and_setup}
We use \texttt{Qwen2.5-Coder} variants (\texttt{7B},\texttt{14B}, and \texttt{32B}) for our experiments due to their strong coding abilities and \texttt{Llama-3.1-8B and Phi-4} for their strong reasoning abilities \cite{hui2024qwen25codertechnicalreport}, skills essential for agents. All models are finetuned using LoRA~\cite{hu2021loralowrankadaptationlarge} (see Appendix \ref{appendix:train_eval_details} for more training and evaluation details). We use the same \react{}~\cite{react} prompt from \citet{appworld-acl24} during inference, but we customize it to be agent-specific (see Appendix~\ref{appendix:prompts}). 
For each variant, we evaluate five different baselines: single-agent (\base{}), and fine-tuned version  (\saft{}), our multi-agent (\mas{}) system proposed in Section \ref{sec:methodology} and its fine-tuned version (\masft{}) and multi-agent system trained with \Algo{} (\masalgo{}). 
We denote agents in multi-agent configurations using the (\orchestrator{}-\exec{}-\critic{}) format, where each placeholder indicates the agent size in billions of parameters (e.g., \texttt{7-7-7}). We also evaluate the multi-agent system with different sized \orchestrator{} models, such as a \texttt{14B} \orchestrator{} with \texttt{7B} \exec{} and \critic{} (\texttt{14-7-7}), and a \texttt{7B} \orchestrator{} with \texttt{14B} \exec{} and \critic{} (\texttt{7-14-14}).
\subsection{Cost-Performance Pareto Analysis}
\label{s:pareto}
\paragraph{Performance} \appworld{} provides two metrics to measure performance: 1) \textbf{Task Goal Completion} (\texttt{TGC}) measures percentage of tasks where agent passes all the human written unit tests for that task. 2) \textbf{Scenario Goal Completion} (\texttt{SGC}) is the percentage of scenarios where the agent passes all the unit tests of each task in that scenario. It measures both consistency and performance across similar tasks.
\vspace*{-1.7mm}
\paragraph{Computational Cost} We use Floating Point Operations (\flops) to measure computational cost\footnote{As we discuss in the limitations section~\autoref{sec:limitations}, we only focus on computational cost and not on memory considerations.}. We approximate \flops per instance as a function of the model parameters and tokens involved in the computation. We use the formula from \citet{kaplan2020scaling} which is $2 \times \#\text{params} \times (\text{input tokens} + \text{output tokens})$.

Our evaluations plot these effectiveness metrics against \flops{} to show the Pareto curves~\cite{yue-etal-2025-masrouter, pimentel-etal-2020-pareto} of different agent setups, thereby providing Pareto-optimal choices at different cost-performance trade-offs.

\section{Results}

\label{sec:results}

\subsection{Effectiveness}
Table \ref{tab:all_results} shows the effectiveness metrics (\texttt{TGC} and \texttt{SGC}) for single and multi-agent setups instantiated with the different size variants of \texttt{Qwen-2.5-Coder} discussed in Section \ref{subsec:model_and_setup}. We observe two findings:


\paragraph{1) Multi-agent systems are more effective than their single-agent counterparts} We expect multi-agent systems with specialized SLMs than their corresponding single agent versions. This is observed in Table \ref{tab:all_results} and \ref{tab:all_results_llama_phi} as the multi-agent system (\mas) is better than the corresponding single-agent baseline (\base) for each parameter variant (e.g., \mas{} 14-14-14 \masalgo{} attains \tgc{} score of $42.3$, while \base{} 14 \saft{} \tgc{} scores $34.5$).
Furthermore, we find that multi-agent systems can even be on par with large single agents (e.g., \texttt{GPT-4o}) in terms of both effectiveness as well as efficiency\footnote{Here we mean efficiency in terms of overall FLOPs and not accounting of engineering overheads that can impact both settings.}. For example, the \texttt{Phi-4 14B-14B-14B} multi-agent gets 46.4, which is close to \texttt{GPT-4o} gets 48.8 on Test-Normal TGC)~\cite{appworld-acl24}.



\paragraph{2) Finetuning with \algo{} creates more effective multi-agent systems} 
To solve \appworld{} tasks, we need models that are good at both complex reasoning as well as coding. Multi-agents powered by code specialist SLMs as base models (e.g., \texttt{Qwen-2.5-Coder})  improve when trained with \algo{}. However, as shown in Table \ref{tab:all_results} and \ref{tab:all_results_llama_phi}, when trained with models that are good at both coding and complex reasoning, such as 
 \texttt{Llama-3.1} and \texttt{Phi-4} base models, we find that \algo{} improves \sgc{} scores on three out of four models, demonstrating its utility for improving robustness as well. Compared to standard fine-tuning, \algo{} yields $18.8\%$ and $18.5\%$ relative gains in \tgc{} scores for the 7B and 14B coding models, respectively. Using \texttt{Llama-3.1} and \texttt{Phi-4}, the relative gains are even larger at $30.8\%$ and $34.5\%$, respectively.

\begin{table}[t!]
\centering
\small
\begin{tabular}{lcc}
\toprule
Model & TGC(\%)& SGC(\%) \\
\midrule
\texttt{Qwen-2.5-Coder-7B} & &  \\
\base{} 7 & 3.6 & 0 \\
\base{} 7 \saft{} & 24.4 & 8.9\\
\mas{} 7-7-7 & 5.9 & 0 \\
\mas{} 7-7-7 \masft{} & 25.6 & 8.9 \\
\mas{} 7-7-7 \masalgo{} & \textbf{30.4} & \textbf{19.6} \\
\midrule
\texttt{Qwen-2.5-Coder-14B} & &  \\
\base{} 14 & 14.9 & 3.6 \\
\base{} 14 \saft{} & 34.5 & 21.4 \\
\mas{} 14-14-14 & 20.2 & 8.9\\
\mas{} 14-14-14 \masft{} & 35.7 & 26.8\\
\mas{} 14-14-14 \masalgo{} & \textbf{42.3} & \textbf{26.8}\\
\midrule
\texttt{Qwen-2.5-Coder-32B} & &  \\
\base{} \texttt{32} & 36.3 & 16.1\\
\mas{} \texttt{32-32-32} & 39.9 & 19.6 \\
\bottomrule
\end{tabular}

\caption{TGC and SGC scores on \appworld{} \texttt{Test-Normal} benchmark for different baselines and our models. We observe that multi-agent systems trained with \algo{} gives best performance across each variant, even better than much base models of much larger variants. SA = Single Agent system, MA = Multi-Agent system, FT = Standard finetune, \algo{} = Finetune with \Algo{}.}
\label{tab:all_results}
\end{table}

\subsection{Pareto Front Comparisons}
\begin{figure}[t!]
  \centering
  \includegraphics[width=\linewidth]{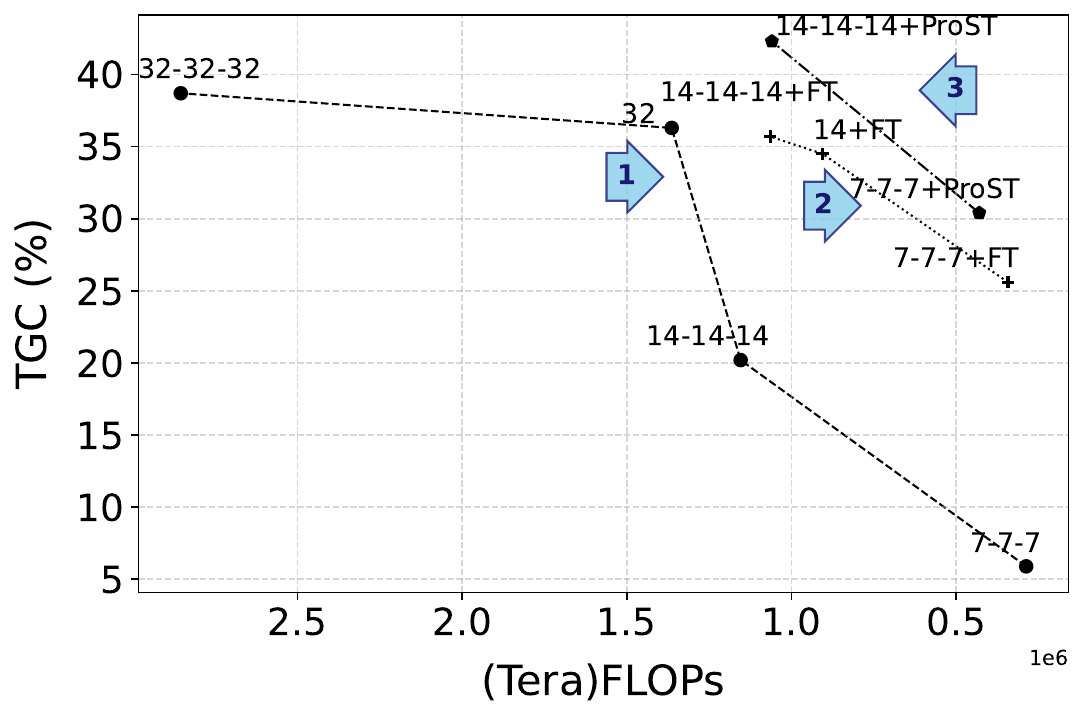}
  \caption{Pareto fronts for three classes of systems: (1) Non-finetuned models, (2) Standard finetuned models, and (3) \loss{}-tuned models.}
  \label{fig:three-pareto}
\end{figure}

Figure \ref{fig:three-pareto} shows the computational efficiency (measured in \flops{}) against task accuracy (\tgc{}) of a subset of our systems using \texttt{Qwen-2.5-Coder} series models. It shows three \emph{Pareto fronts} for non-finetuned, finetuned, and \algo{} fine-tuned models. The figure supports three key findings:
\paragraph{1) \algo{} improves Pareto optimality of multi-agent systems } Pareto-front is worst for non-finetuned models. Fine-tuning gives more Pareto-optimal models, observed by Pareto fronts (2) and (3), shifting rightwards, i.e., reduces computational cost while improving effectiveness. 
However, models trained using \algo{} have the best tradeoff between accuracy and computational efficiency. This shows the effectiveness of \algo{} in training more effective models with lower computation costs.
\paragraph{2) Optimality is not dependent on total parameter size alone} One would expect a system with a larger number of overall parameters to have a higher computational cost and to perform better in general. However, when comparing the non-fine-tuned \texttt{32B} single-agent setup to the \texttt{14-14-14} \algo{} multi-agent setup, we observe that the multi-agent system has more parameters (\texttt{32B} vs \texttt{42B}), yet computational cost is lower ($1.36$ vs $1.06$ TeraFLOPs). This is due to the overall computational cost depending on the overall length of trajectories produced by the different systems, which in turn also depends to some degree on their effectiveness. Notably, the fine-tuned 1\texttt{4-14-14} \algo{} multi-agent setup achieves a higher score than both the non-fine-tuned \texttt{32B} single-agent setup and the much larger \texttt{32-32-32} multi-agent setup (\texttt{96B} parameters), highlighting that effectiveness is not solely 
determined by total parameter size.

\begin{figure}[t]
    \centering
    \includegraphics[width=\linewidth]{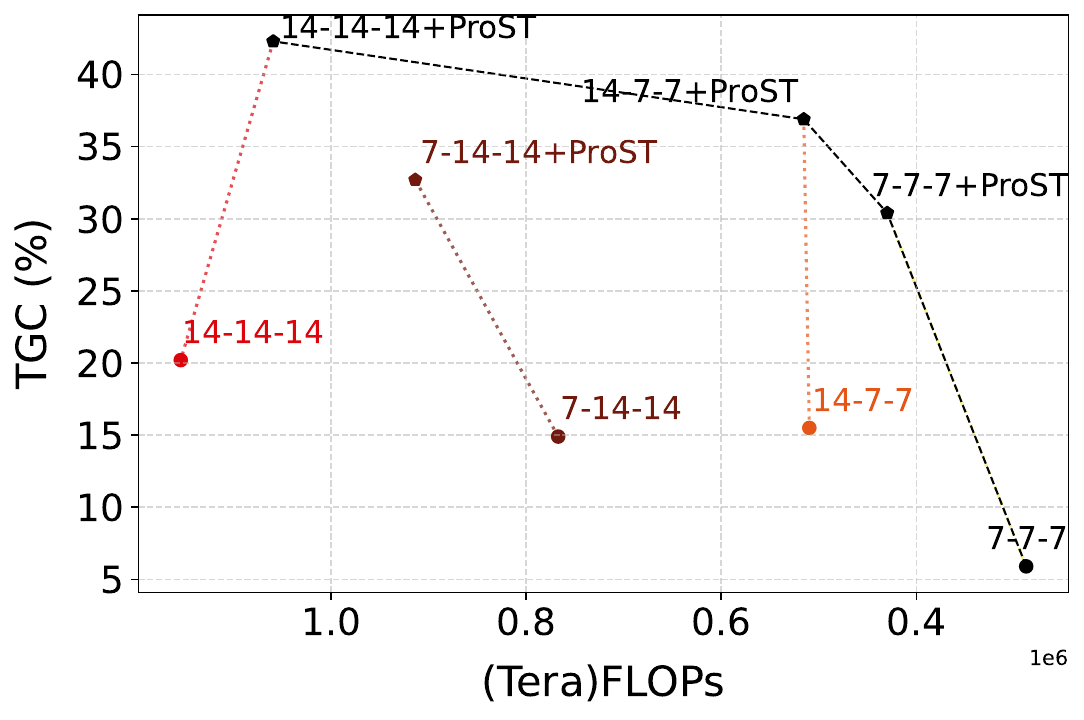}
    \caption{Stronger \orchestrator{s} trained by \Algo{} achieve better Pareto optimality in terms of computation and accuracy. Points closer to the top-right indicate better trade-offs.}
    \label{fig:big-orch}
\end{figure}

\paragraph{3) Stronger \orchestrator{} is a more optimal choice} 
Recent work shows that powerful planning/\orchestrator{} agents can improve task effectiveness~\citep{erdogan2025planandactimprovingplanningagents}. We study whether allocating higher capacity to the \orchestrator{} leads to more Pareto-optimal solutions with \algo{} (see \autoref{fig:big-orch}.) To this end, we create two system variants: (i) \texttt{7-14-14}, which uses a weaker \orchestrator{} (\texttt{7B}), and (ii) \texttt{14-7-7}, which uses a stronger \orchestrator{} (\texttt{14B}).

We find that multi-agent systems with a strong \orchestrator{} are more optimal than ones with a weak \orchestrator{} of the same parameter count (see Table~\ref{tab:appendix_results_different_models_mas} in Appendix). As shown in the figure, \texttt{14-7-7 + \algo{}} has higher \tgc{} and is also more efficient than \texttt{7-14-14 + \algo{}}. Also, \algo{} training reduces the overall \flops{} for all MAS systems with strong \orchestrator{}s (i.e. \texttt{14B}) while improving effectiveness, when compared to their non-fine-tuned versions. Whereas, \algo{} training a weak \orchestrator{} (i.e. \texttt{7B}) improves effectiveness but does not reduce overall \flops{}. This is likely because poor orchestration (i.e. planning and guidance) causes more trial and error before success, leading to longer trajectories.

\section{Analysis}

\paragraph{Importance of progressive subtask selection} 
\algo{} can be seen as training for a subset of tasks in each epoch using a specific progressive inclusion strategy. 
Here we test the necessity of the subset selection by comparing against including all subsets of tasks in each epoch (\texttt{ALL}), and the necessity of our specific progressive strategy by comparing against both (i) random subset selection at each epoch (\texttt{Random})\footnote{Note that this strategy is constructed to ensure that all subtasks are shown to the model during the training. See Appendix \ref{appendix:random_subtask_selection} for details.}, and (ii) an inverse strategy that includes all subtasks (full task) first, and removes subtasks as training progresses(\texttt{Decrement}).

\begin{figure*}[t!]
  \centering
  \includegraphics[width=\linewidth]{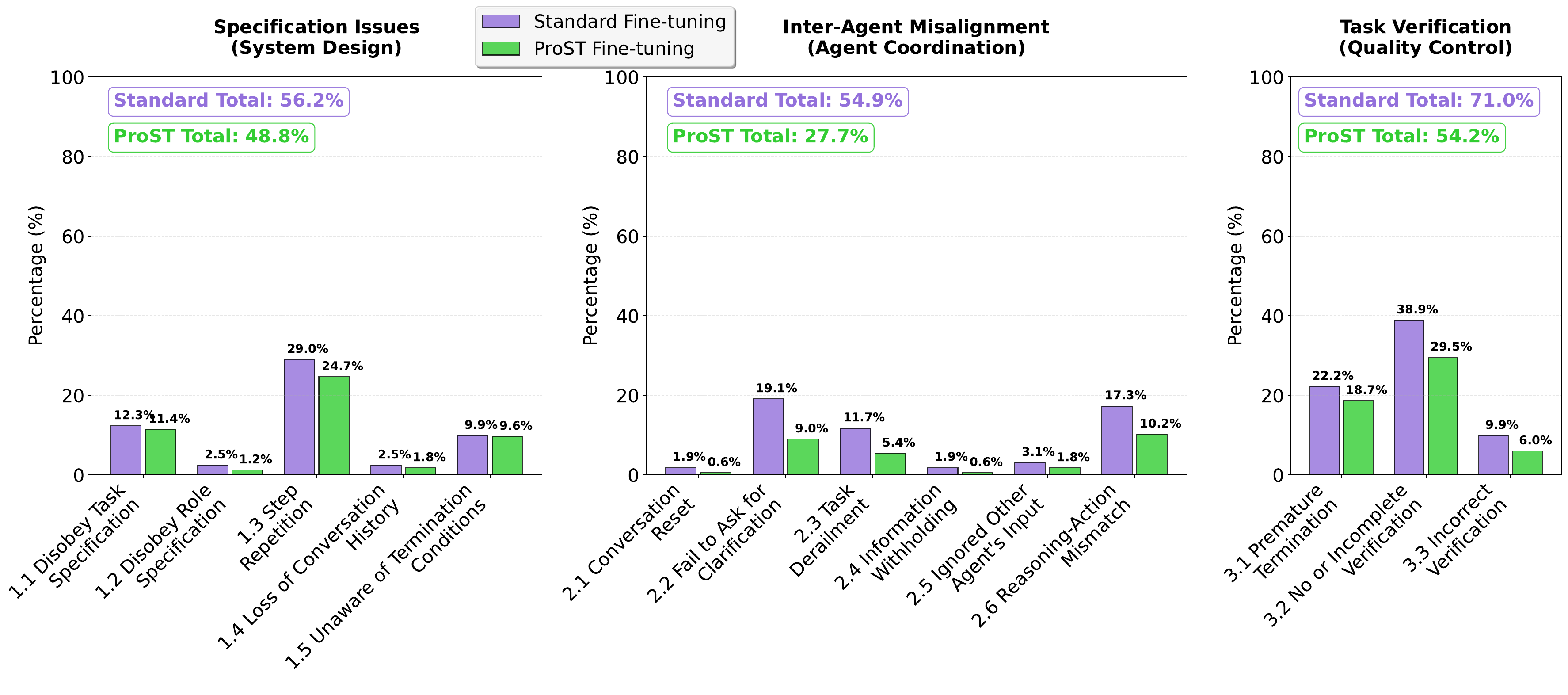}
    \caption{Failure mode analysis using the MAST~\cite{cemri2025multiagentllmsystemsfail} taxonomy (Specification Issues, Inter-Agent Misalignment, Task Verification) and the LLM-as-a-Judge pipeline (\texttt{Llama-3.3 70B} as the judge). \Algo{} (\algo{}) consistently reduces error modes across all categories.}
  \label{fig:mast_ft_vs_prost}
\vspace{-1em}
\end{figure*}

\begin{figure}[t!]
  \centering
  \includegraphics[width=\linewidth]{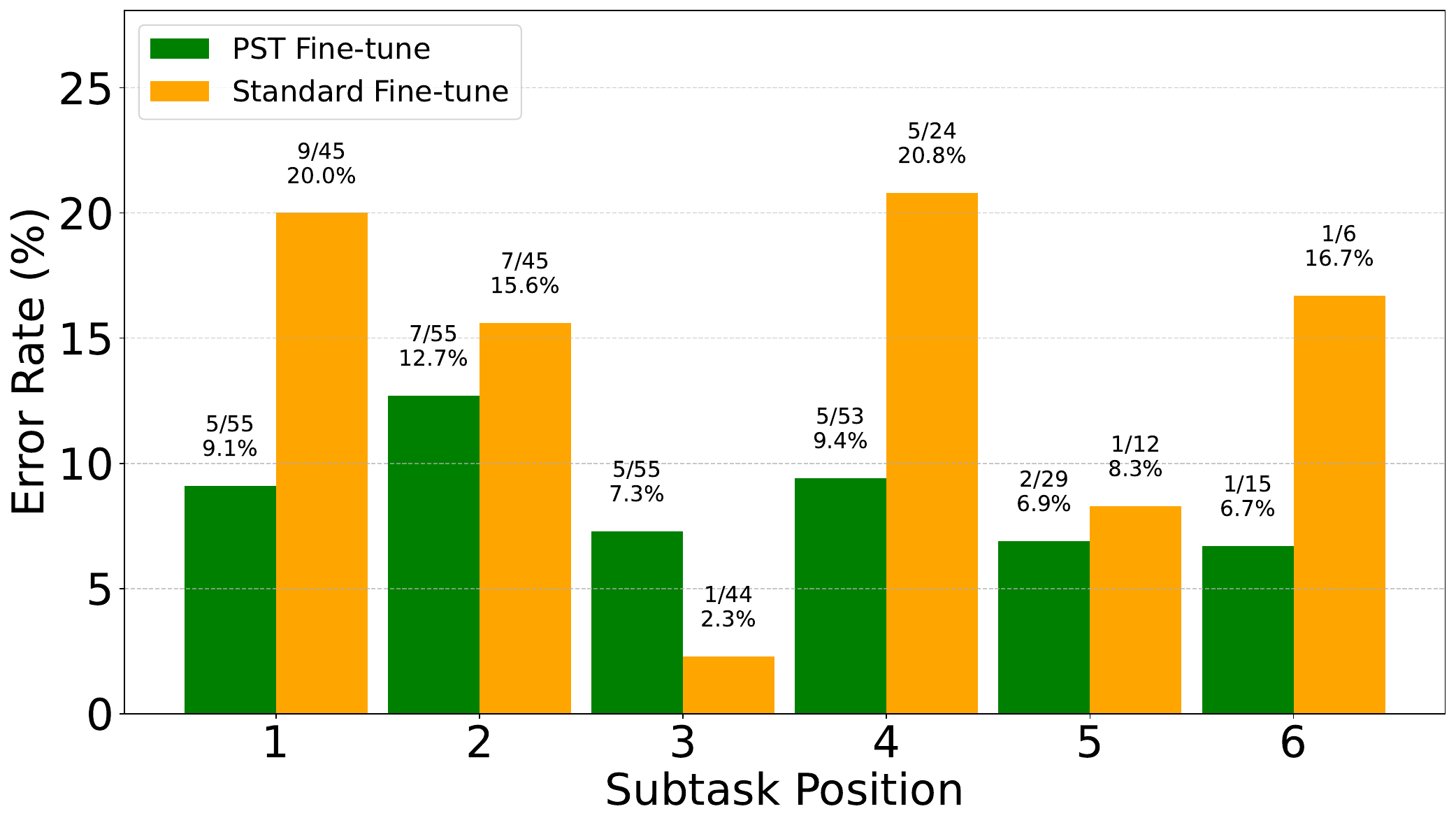}
    \caption{Error rates per subtask position for \algo{} Fine-tuning (green) and Standard Fine-tuning (orange), computed only on successful tasks. The error rate is defined as (number of successful tasks with at least one error at subtask $i$ / total number of successful tasks that reached subtask $i$) $\times$ 100\%. We consider only the subtasks which has more than 5 successful tasks from both settings. \algo{} Fine-tuning shows lower error rates across most subtasks, while Standard Fine-tuning exhibits significantly higher error rates.}
  \label{fig:error_rate_in_subtasks}
\vspace{-1em}
\end{figure}

\begin{table}[tb!]
\centering
\small
\begin{tabular}{lcc}
\toprule
Selection Strategies & \tgc{} & \sgc{}\\
\midrule
\small \texttt{Qwen-2.5-Coder-14B MAS [\algo{}]} & & \\
- \texttt{All} & 35.7 & 26.8 \\ 
- \texttt{Random} & 34.5 & 23.2 \\
- \texttt{Decrement} & 39.9 & 26.8 \\
- \texttt{Ours} & \textbf{42.3} & \textbf{26.8} \\
\bottomrule
\end{tabular}
\caption{Ablation: Effect of different subtask selection strategies in \algo{}.}
\label{tab:pst_configurations}
\vspace{-1em}
\end{table}
Table \ref{tab:pst_configurations} shows that random selection yields the worst performance, whereas progressively learning more subtasks either via (\texttt{Decrement} or \texttt{Ours}) provides gains over training on all subtasks from the start. However, \texttt{Ours}, the progressive curriculum whereby we learn the task-specific subtasks first, is still substantially better than its inverse \texttt{Decrement}, likely because it better reflects the natural dependencies between the subtasks. However, we see that \sgc{} scores are the same for \texttt{All}, \texttt{Decrement}, and \texttt{Ours}, showing that selection has no affect on solving tasks across same scenarios.

\paragraph{\loss{} reduces error rates in subtasks and MAST taxonomy} 
One of the main motivations for \algo{} was our observation that SLMs are unable to learn all subtasks effectively under standard fine-tuning. \algo{} aims to address this by training models to incrementally learn more subtasks. Figure \ref{fig:error_rate_in_subtasks} compares the rates of error at each subtask position from successful completion during inference for standard finetuning and \algo{} by counting the number of error messages returned by the \appworld{} environment. We find that \algo{} has a lower error rate than standard fine-tuning at most subtask positions. This shows that, with \algo{}, SLMs learn subtasks more effectively. 

\begin{figure}[!t]
    \centering
    \includegraphics[width=\linewidth]{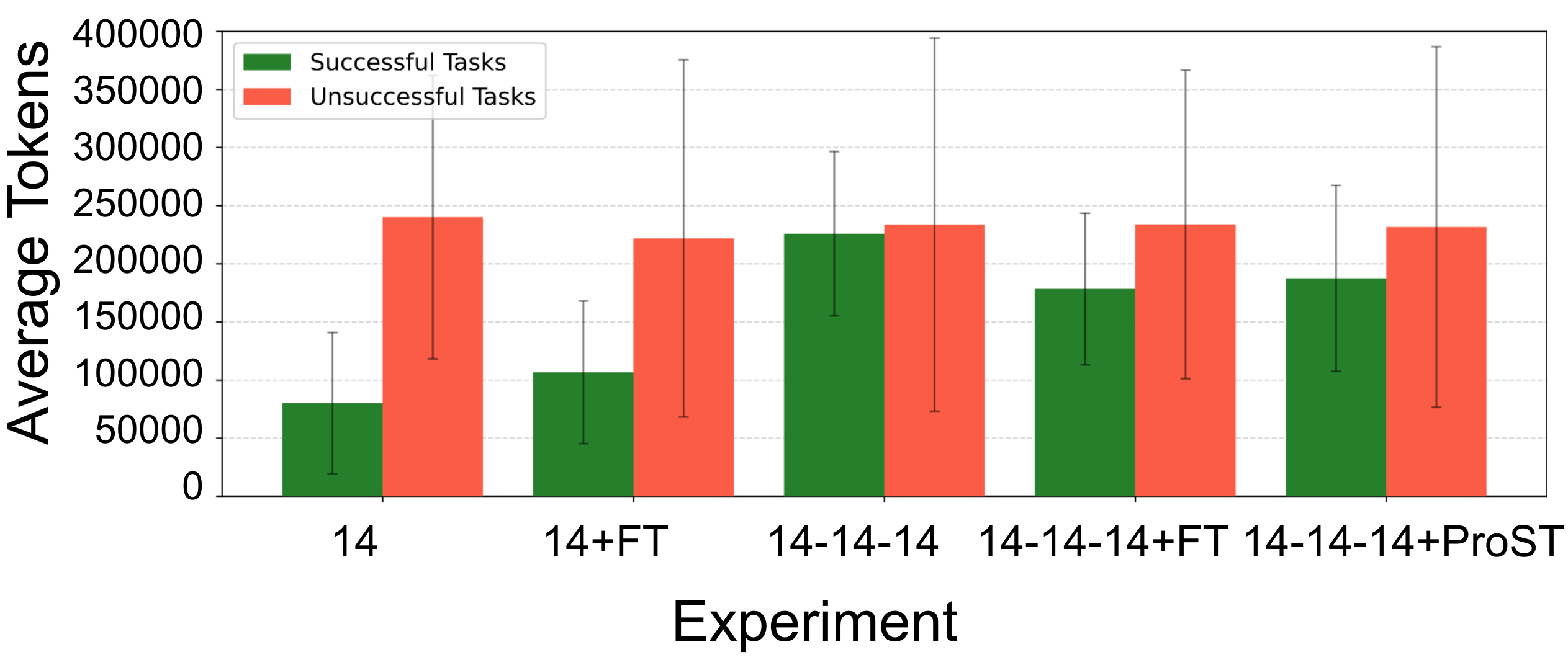}
    \caption{Comparison of total tokens across successful and unsuccessful task completion. Green bars indicate tokens processed in successful tasks, and red bars represent tokens processed in failed tasks. Percentage labels above each bar show the ratio of processed tokens that resulted in completing tasks.}    
    \label{fig:success_v_unsuccess}
\end{figure}

To analyze the errors further, we adopt MAST (Multi-Agent System Failure
Taxonomy) \cite{cemri2025multiagentllmsystemsfail}, an empirically grounded taxonomy for categorizing errors in multi-agent systems, which categorizes failures into \textit{Specification Issues}, \textit{Inter-Agent Misalignment}, and \textit{Task Verification}. We use their validated LLM-as-a-Judge pipeline (with \texttt{Llama-3.3 70B} as the judge) to classify errors across these categories. As shown in Figure \ref{fig:mast_ft_vs_prost}, \Algo{} (\algo{}) consistently reduces error rates across all categories and subcategories, with highest improvements in inter-agent coordination failures.
\vspace*{-1.7mm}
\paragraph{How do trajectory lengths compare?} 
We compare the trajectory lengths of single and multi-agent systems. Figure \ref{fig:success_v_unsuccess} shows the distribution of tokens for the \texttt{14B} based systems in successfully completed test set tasks compared to the unsuccessful ones (see Figure~\ref{fig:tokens_count_all} in the Appendix for the full results across 7B, 14B, and 32B models) . The trajectory lengths, i.e, the average number of tokens, for unsuccessful tasks is higher than for successful ones for all systems with higher variance. Unsuccessful tasks are likely to have more trial-and-error behavior and more errors, which add to the trajectory length. Multi-agent systems have longer trajectories for successful tasks compared to single-agent systems in part because of the overheads involved in separating the roles and the resultant communication. Fine-tuning, standard, and \algo{} have shorter trajectories for successful tasks, likely due to reduced subtask error rates.

\begin{table}[t!]
\centering
\small
\begin{tabular}{lcc}
\toprule
Model & TGC(\%)& SGC(\%) \\
\midrule
\texttt{Llama-3.1-8B} & &  \\
\base{} 8 & 6.6 & 1.8 \\
\base{} 8 \saft{} & 26.2 & 8.9\\
\mas{} 8-8-8 & 5.9 & 0 \\
\mas{} 8-8-8 \masft{} & 25 & 7.1 \\
\mas{} 8-8-8 \masalgo{} & \textbf{32.7} & \textbf{19.6} \\
\midrule
\texttt{Phi-4-14B} & &  \\
\base{} 14 & 15.5 & 3.6 \\
\base{} 14 \saft{} & 32.1 & 14.3 \\
\mas{} 14-14-14 & 30.9 & 14.3\\
\mas{} 14-14-14 \masft{} & 34.5 & 19.6\\
\mas{} 14-14-14 \masalgo{} & \textbf{46.4} & \textbf{28.6}\\
\bottomrule
\end{tabular}

\caption{TGC and SGC scores on \appworld{} \texttt{Test-Normal} benchmark for different baselines and models (Llama-3.1-8B and Phi-4-14B). We notice similar performance tradition between Qwen series models and Llama-3.1-8B \& Phi-4-14B. We observe that multi-agent systems trained with \algo{} have the best performance in all settings. SA = Single Agent system, MA = Multi-Agent system, FT = Standard finetune, \algo{} = Finetune with \Algo{}.}
\label{tab:all_results_llama_phi}
\vspace{-1em}
\end{table}

\paragraph{Does \algo{} generalize to models of different capabilities?}
We also used \algo{} to train two other non-coding models with different generative capabilities, \texttt{Llama-3.1-8B} and \texttt{Phi-4-14B} and evaluated them on both \texttt{Test Normal} (see Table~\ref{tab:all_results_llama_phi}) and \texttt{Test Challenge} (see Table~\ref{tab:test_C_results}) sets. For \texttt{Test Normal}, Llama-3.1-8B and Phi-4-14B yield significant performance improvements. Phi-4-14B performed even better than Qwen-2.5-Coder-14B. Phi-4 \texttt{14-14-14 \masalgo{}} shows $46.4\%$ and $28.6\%$ in TGC and SGC in Test Normal, respectively, and is the highest among the different sized models. 

\paragraph{Can \algo{} generalize to out-of-distribution data (OOD)?}
\appworld{} also provides an out-of-distribution \texttt{Test-Challenge} test set, with tasks requiring more API calls and use at least one unseen application.
We evaluate \algo{} on this test set, presenting results for \texttt{Qwen-2.5 Coder 14B} and \texttt{Phi-4 14B} in Table \ref{tab:test_C_results}.
We find that \texttt{Phi-4-14B} with \algo{} outperforms other fine-tuning and multi-agent baselines with $17.8$\% in \tgc{} and $8.6$\% in \sgc{}. While \texttt{Qwen-2.5-Coder-14B}'s fine-tuned single-agent achieves the best performance in its model family, its \algo{} variant has higher accuracy than the corresponding fine-tuned multi-agent setup. 
These results show that \algo{}, with appropriate base models, can help improve  OOD generalization. 
More results for this test set are present in Appendix Table~\ref{tab:test_C_results_7b_8b}.

\begin{table}[tb!]
\centering
\small
\begin{tabular}{lcc}
\toprule
Model & TGC & SGC \\

\midrule
\texttt{Qwen-2.5-Coder-14B} & &  \\
\base{} 14 & 4.3 & 2.2 \\
\base{} 14 \saft{} & \textbf{16.8} & \textbf{5.8} \\
\mas{} 14-14-14 & 10.1 & 3.6 \\
\mas{} 14-14-14 \masft{} & 13.7 & 5.0 \\
\mas{} 14-14-14 \masalgo{} & 14.1 & \textbf{5.8} \\

\midrule
\texttt{Phi-4-14B} & &  \\
\base{} 14 & 6.2 & 2.2 \\
\base{} 14 \saft{} & 14.1 & 3.6 \\
\mas{} 14-14-14 & 12.2 & 4.3 \\
\mas{} 14-14-14 \masft{} & 12.5 & 7.2 \\
\mas{} 14-14-14 \masalgo{} & \textbf{17.8} & \textbf{8.6} \\

\bottomrule
\end{tabular}
\caption{Performance comparison in single-agent and multi-agent settings on \emph{Test Challenge} (Test-C) set for  \texttt{Qwen-2.5-Coder-14B} and \texttt{Phi-4-14B}.}
\label{tab:test_C_results}
\vspace{-1em}
\end{table}
\vspace*{-1.7mm}
\section{Conclusion}
\vspace*{-2mm}
SLM based multi-agents present  a viable alternative for solving complex problems. However, the limited capacities of SLMs prevent them from learning all subtasks in long trajectory problems. In this work, we introduced a new \Algo{} algorithm that helps address this challenge. Progressively introducing subtasks during training allows the limited capacity SLMs to more effectively learn the subtasks resulting in improved overall effectiveness. To better understand the cost-effectiveness trade-offs between single and multi-agent setups, we conduct pareto analyses which yield a holistic view of the performance of these systems. Our experiments on \appworld{}, a popular agentic benchmark demonstrates the utility of \algo{} for addressing complex problems. 


\section*{Limitations}
\label{sec:limitations}

\paragraph{Generalizability} We use \appworld{} as a test bed for complex reasoning problems. While our design for the agents i.e., their roles, and \algo{} training paradigm are broadly applicable to many agentic frameworks, and is not \appworld{} specific (detailed explanation in Appendix \ref{appendix:prost_generalizability}), some aspects of our modeling and agentic design are influenced by specifics of \appworld{}. For example, we showed through ablation studies that the order of introducing subtasks in \algo{} does indeed matter. However, the choice of introducing the first and last subtasks at the end of training is specific to \appworld{}, and it may not generalize well to other benchmarks. 

\paragraph{Training Data} The training dataset is LLM generated. While we only use successful trajectories, the overall quality of these trajectories (e.g., diversity) has not been evaluated. Also, some parts of the multi-agent training instances are created using a closed-source LLM (\texttt{Claude-3.7-Sonnet}). LLM generation of synthetic data using frontier (non open-source) models, while widely used \citep{saqib2025teaching, wang2025critique, li2023synthetic, tang2023does}, has its limitations in terms of reproducibility. We will release these instances to ensure replicability and comparisons against other methods for training agents. However, the standard issues of using closed-source LLM will make it difficult to compare against other means of generating training data, unless the method is replicated. We will release code and data to support replicability as much as possible.

\paragraph{Inference-time Computation Cost} \Algo{} optimizes agents in multi-agent architectures such that models perform better at similar inference time cost, measured in \flops{}. However, our analysis does not include memory considerations. This can be a significant bottleneck as we load multiple instantiations of SLMs, leading to a large GPU overhead. For instance, when loading the \texttt{14-14-14} multi-agent system, we need one H100 GPU (80GB each) for each agent, totaling around 240GB of memory. While parameter count is taken into account during inference computation cost, we do not account actual cost of loading the whole. We thus carefully scope our claims to only runtime costs as measured by \flops{}.

\paragraph{Out-of-domain Aspects:} Since we finetune models on training data with limited scope, models may have learned short-cuts, or overfit to the distribution of the training data. The mixed results on out-of-distribution test data in table \ref{tab:test_C_results} and \ref{tab:test_C_results_7b_8b} and  suggests scope for future work.

\section*{Acknowledgements}

This work is supported in part by a SUNY-IBM Artificial Intelligence Collaborative Research Alliance grant, an Amazon Research Award 2023, and a NSF NAIRR compute credits grant \#240140. Any opinions, findings, and conclusions or recommendations expressed in this material are those of the author(s) and do not reflect the views of Amazon.

\bibliography{anthology,custom}

\bibliographystyle{acl_natbib}
\newpage
\appendix

\section{Data Preparation and Trajectory Transformation}
\subsection{Single-Agent Trajectory Generation}
\label{appendix:single-agent-data-creation}

\appworld{}~\cite{appworld-acl24} provides training pairs consisting of input prompts and output states. Specifically, it offers 90 training and 57 development tasks, each with its corresponding output state (i.e., Python solution). We use all 147 tasks to create a fine-tuning dataset for both single-agent and multi-agent setups.
To increase the dataset size and introduce solution diversity for each task, we use \texttt{Llama-3.3-70B-Instruct}~\cite{llama2025instruct33} as the Re-Act agent across 20 different temperature settings. We start at a temperature of $0.05$ and increment by $0.05$ each time, up to $1.0$. To reflect practical scenarios, we first attempt to solve all $147 \times 20 = 2,940$ tasks without providing the output state to the agent. Our base \texttt{Llama-3.3-70B-Instruct} model successfully solves 1,034 out of 2,940 candidate trajectories across the different temperature settings.
For the remaining tasks, we provide abstract instructions instead of direct code. We generate these instructions in natural language using \texttt{Llama-3.3-70B-Instruct}. By feeding these abstract instructions to the agent for the 1,906 unsolved candidates, we obtain 1,396 new gold trajectories. For the final 510 candidates, we convert the Python code solutions into pseudocode using \texttt{Llama-3.3-70B-Instruct}. We avoid giving direct Python code as hints because the agent tends to reproduce the code verbatim. Assisting in pseudocode format yields 414 additional gold trajectories.
In total, we collect 2,844 gold trajectories. This approach ensures diverse solutions for each task across different temperature settings. When our agent fails to solve tasks in the first phase (i.e., limit exceeds), we use our curated abstract assistance. We did not generate new training tasks for \appworld{}, since creating and validating new tasks requires specific environment support, which is challenging to set up.

\subsection{Transformation to Multi-Agent Trajectories}
\label{appendix:sat-to-mat}
We ensure that each agent in the multi-agent system is trained effectively to handle its specific role while maintaining overall system coherence. To achieve this, we have two options at this point to prepare agent-specific training data. One option is direct distillation. First, we use \texttt{Llama-3.3-70B-Instruct} for this purpose. However, open-source models such as \texttt{Llama-3.3-70B-Instruct} are not trained on such long-horizon tasks for multi-agent setting, leading to fragile plans and overall less informative trajectories. On the otherhand, direct distillation of multi-agent trajectories from frontier models is both prohibitively expensive and has fundamental technical limitations. The costs of direct distillation exceed well beyond typical research budgets—our estimate worked out to between $20-25,000$ USD just to obtain the trajectories. To obtain around 3000 trajectories we will have to get around 12000 sample generations from the frontier model. With each generation costing around $2$, this works out to $24,000$ dollars. 


In contrast, our cost-efficient trajectory transformation approach leverages existing successful single-agent solutions as a foundation, then systematically decomposes them into coordinated multi-agent workflows. We leverage in-context learning capabilities of LLMs by providing the teacher LLM (\texttt{Claude-3.7-Sonnet})~\citep{anthropic2025claude37} with successful single-agent trajectories and prompting it to decompose them into structured multi-agent orchestrations. Given a single-agent trajectory that successfully completes a task in the AppWorld environment, we prompt the LLM to partition the sequence of actions into logical subtasks while preserving the exact execution steps that led to success. Our transformation approach maintains a critical constraint: each executor step (both the reasoning and code) must be identical to the original single-agent trajectory steps. Crucially, we do not allow the model to invent new code or modify existing execution logic - it can only redistribute the original successful steps across subtasks and add exit commands to signal subtask completion. The teacher LLM successfully produces valid 2,708 trajectories from 2844 single agent trajectories. This approach ensures that the generated multi-agent trajectories align with the real execution environment and we further validated these transform actions again by running into the AppWorld environment. This approach proved both economically feasible (cost was only $300$ USD) and effective for our purposes. 

\begin{figure}[t]
  \centering
  \includegraphics[width=\linewidth]{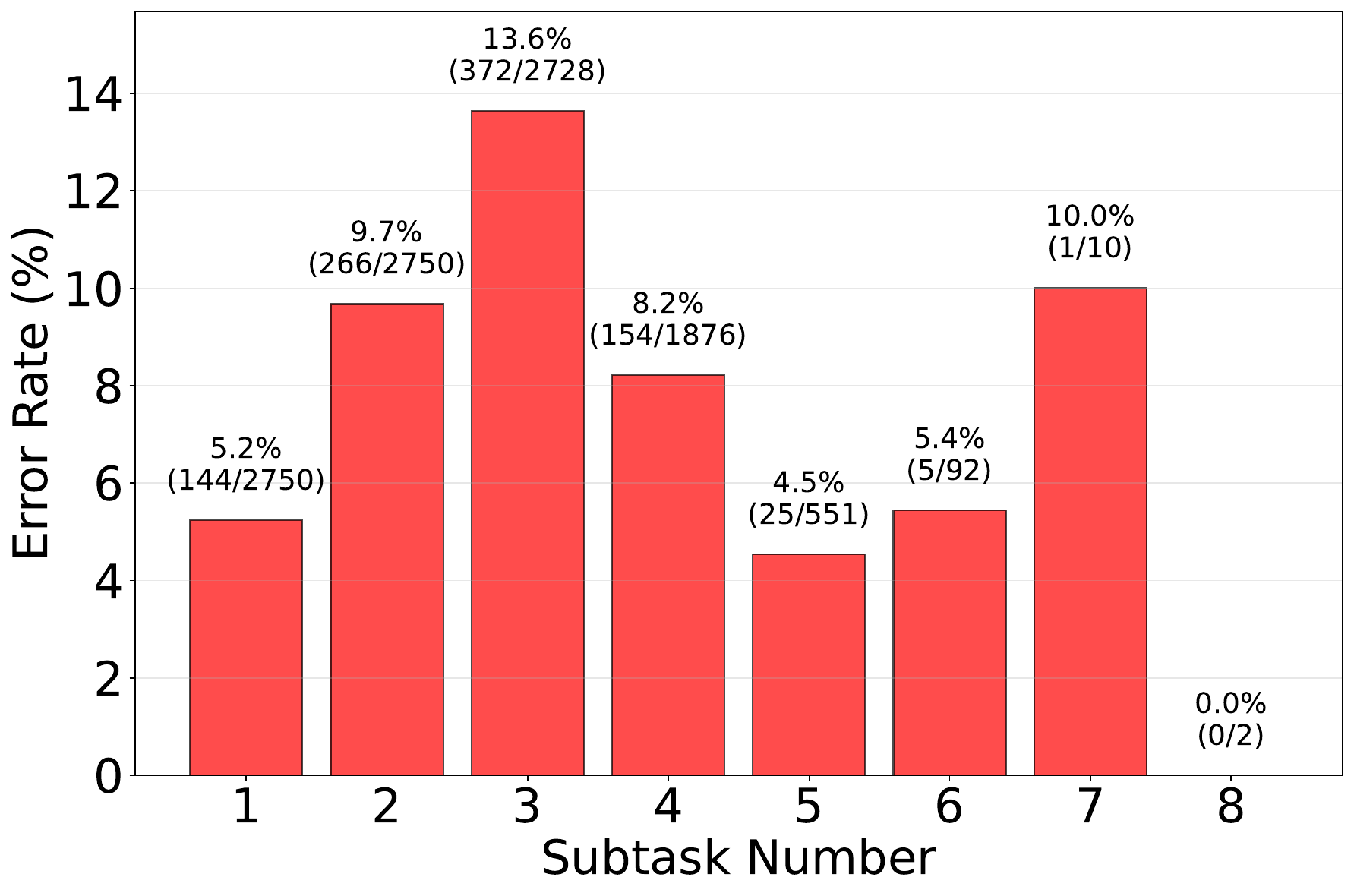}
  \caption{Error rate percentage for each subtask position across all training tasks. It shows what percentage of tasks have at least one error in each subtask position (1st subtask, 2nd subtask, 3rd subtask, etc.). We define the error rate function as (number of tasks that have at least one error in $i^{th}$ subtask position / total number of tasks that have $i^{th}$ subtask position ) $\times$ 100\%}
  \label{fig:error_rate_in_train_subtasks}
\end{figure}

While the impact of different distillation strategies can add useful knowledge this is somewhat orthogonal to our key contributions of pareto comparisons of single/multiagent systems, improving multiagents using SLMs through ProST training. The idea of using synthetically distilled dataset is a part of our work but not our main focus. We further analyze error rates for each subtask of the multi-agent trajectories (depict in Figure \ref{fig:error_rate_in_train_subtasks}). It shows our dataset covers error correction patterns across all subtasks. We finetune specialized agents of different sizes (7B and 14B), as well as single agents as baselines. Agents in the multi-agent systems are trained on their respective datasets in two ways: (1) standard supervised finetuning, and (2) \Algo{} (\algo{}) (details in Section \ref{sec:pst}.)


\subsection{Pipeline Rationale and Verification}
The primary reason we used the specific pipeline of \texttt{Llama-3.3-70B} to generate single-agent trajectories and frontier model (i.e., \texttt{Claude-3.7-Sonnet}) to translate into multi-agent trajectories is cost. While our approach enables cost-effective scaling, it also introduces variability. These imperfections can affect fairness by biasing agents toward particular solution styles, and they limit reproducibility since other researchers may obtain different trajectories if different LLMs or prompts are used. To mitigate this, we will release all generated trajectories and prompts, enabling others to train under identical conditions. Understanding how the LLMs structure reasoning in natural language is an interesting study that can shed further light on different reasoning patterns, but one that is beyond the scope and goals of our work.

Notably, both the single-agent and the multi-agent trajectories are fully verified. They are gold trajectories in the sense that using them will solve the corresponding tasks. Both versions share a bulk of their trajectories, with the added parts in multi-agent trajectories being mostly content specific to multi-agent aspects of the problem. Our generated trajectories are fully verified --- they are actually validated via execution in the AppWorld environment. We only retain trajectories that actually work for the given task. We conducted further sanity checks on a limited set for prompt engineering to ensure that there are no formatting or structural errors and that the steps remain consistent with the intended task goals. There were no errors in the sub-task goals or the description, although we found some cases where the description can be improved in small ways. The reason for the low-levels of noise here is that the original single agent trajectories we use are gold trajectories -- actually are verified. Together, this combination of automatic environment-based validation and manual spot-checking shows that our dataset is both high-quality and task-valid.

We use heterogeneous teachers because generating high-quality action traces and plan-level abstractions requires models with different strengths. This approach mirrors~\citet{erdogan2025planandactimprovingplanningagents}, which similarly separates plan and act supervision, employs open-source models for execution traces, and frontier models for plan generation and expansion. Both approaches use synthetic data generation, annotating ground-truth trajectories with feasible plans to strengthen the planner – a method shown to improve planner quality in. Fairness is further preserved by applying the same evaluation protocols and training pipeline components where applicable across settings, leaving the agent paradigm (single-agent versus multi-agent) and the multi-agent specific learning method (\algo{}) as the sole intended variables. As reflected by consistent Pareto and \tgc{}/\sgc{} gains across model scales, our observed effect comes from \algo{} and role specialization, not from differences in teacher models.

\section{Implementation Details}
\subsection{Training and Evaluation Configuration}
\label{appendix:train_eval_details}
Our base models are finetuned with Low-Rank Adaptation (LORA)~\cite{hu2021loralowrankadaptationlarge}. We use rank (r=16), alpha ($\alpha$=32) and dropout = 0.05. We use a learning rate of 2.0e-4 with a cosine scheduler. To stabilize training, we set the gradient clipping to 1.0 and the weight decay to 0.01. We train the model for 5 epochs. We use a warm-up ratio of 0.1. The maximum sequence length is set to 20,480 tokens, as our training trajectories are long. All of our experiments are conducted on 4 NVIDIA H100 GPUs with 80GB of memory each. We use Accelerate integrated with DeepSpeed (ZeRO-3) for multi-GPU training. This setup provides efficient memory management and optimization. For both training and evaluation, each agent took ~8 GPU hours.
For evaluation, we use vLLM~\cite{kwon2023efficientmemorymanagementlarge} servers. For both single-agent and multi-agent setups, we perform inference at T=0.1 and T=0.2 for \textit{Test Normal} set and \textit{Test Challenge}, respectively. We set the maximum sequence length to 65536 and enable prefix caching. For the single-agent setting, we allow up to 50 turns for evaluation. For the multi-agent, \orchestrator{} can plan up to 12 subtasks, and each time \exec{} allows max 15 turns (for the \textit{Test Challenge} set, we set max 20 turns) to complete each subtask. A task is considered failed if an agent calls the task completion API with a fail status or doesn't complete the task within the maximum allotted limits.

\subsection{Random Subtask Selection Strategy}
\label{appendix:random_subtask_selection}
To create training data for progressively random subtask(s) selection finetuning strategy, we employ a randomized epoch assignment mechanism to ensure comprehensive exposure of all subtasks during model training. For each subtask within a task, we generate two random integers between 0 and 4, then create an inclusive range of epochs from the minimum to the maximum value. This approach guarantees that every subtask appears in at least one epoch (when both random numbers are the same) and can span multiple consecutive epochs (when the numbers differ). This strategy ensures that all subtasks receive adequate exposure throughout the 5-epoch training process. 

\subsection{\algo{} Details}
\label{appendix:prost_details}
We always start with at least 2 subtasks. We progressively add task-specific subtasks, primarily during epochs 0--2, and include the first subtask (login in \appworld{}) and task completion subtasks in epochs 3--4 if the total number of subtasks is at most 5. If there are 6 or more subtasks, we place greater emphasis on task-specific subtasks in epochs 0--3 and add non-task-specific subtasks in epoch 4. In general, we do not strictly assign subtasks to specific epochs; instead, our approach adds subtasks when there are enough available subtasks to add at each epoch. Otherwise, we add subtasks after every $k$ epochs (e.g., $k=2,3$). For each task, we make sure that all subtasks are seen during the full $5$-epoch training.

\section{Appworld Details}
\label{appendix:appworld_details}
\appworld{} \cite{appworld-acl24} is a framework designed to assess the ability of autonomous agents in solving real-world problems by interacting with an environment of everyday applications. The framework provides the user with two components: the \textbf{\appworld{} Execution Engine} and \textbf{\appworld{} Benchmark}. 

\appworld{} Execution Engine provides an execution environment where agents can interact with abstract versions of the following nine commonly used applications: Amazon, Spotify, Venmo, Gmail, Todoist, SimpleNote, Splitwise, Phone, and FileSystem. These applications are simulated using $457$ different APIs and $101$ database tables with $~370K$ rows that represent around $100$ users. The execution engine is designed to create a safe and well-tested (98\% coverage with $1,780$ unit tests) environment for testing agents' ability to interact with applications by writing code to invoke APIs. 

\appworld{} Benchmark provides a dataset consisting of $750$ realistic task instructions across $250$ different scenarios of app usage in the \appworld{} engine. The tasks are divided into $105$ train-set $60$ validation set\footnote{We found only 90 train and 57 dev tasks are available in the appworld when we load the datasets.}, $168$ in-distribution \emph{Test Normal} test set, and $417$ out-of-distribution \emph{Test Challenge} testbed with unseen apps and APIs. The tasks are created to utilize, on average, $1.8$ apps or make $9.8$ API calls by writing $\sim 50$ lines of code. The task is evaluated using \textbf{State-Based Programmatic Evaluation}, where the final state of the database is checked against several manually written assertions that must pass if the task was completed correctly.

\begin{table}[ht]
\centering
\small
\begin{tabular}{lcc}
\toprule
Model & TGC(\%)& SGC(\%) \\
\midrule
\texttt{Qwen-2.5-Coder} & &  \\
\mas{} 14-7-7 & 15.5& 3.6 \\
\mas{} 7-14-14 & 14.9 & 3.6 \\
\mas{} 14-7-7 \masft{} & 30.4& 17.9\\
\mas{} 7-14-14 \masft{} & 26.8& 12.5 \\
\mas{} 14-7-7 \masalgo{} & 36.9&  26.8\\
\mas{} 7-14-14  \masalgo{} & 32.7& 16.1\\
\bottomrule
\end{tabular}

\caption{Extended results table with different model sizes multi agent system. TGC and SGC scores on \appworld{} \texttt{Test-Normal} benchmark for different baselines and our models. MA = Multi-Agent system, FT = Standard finetune, \algo{} = Finetune with \Algo{}.}
\label{tab:appendix_results_different_models_mas}
\end{table}

\begin{table}[t!]
\centering
\small
\begin{tabular}{lcc}
\toprule
Model & TGC(\%)& SGC(\%) \\
\midrule
\texttt{Qwen-2.5-Coder-7B} & &  \\
\base{} 7 & 0.2 & 0.0 \\
\base{} 7 \saft{} & 8.9 & 2.9 \\
\mas{} 7-7-7 & 1.7 & 0.0 \\
\mas{} 7-7-7 \masft{} & 7.2 & 3.6 \\
\mas{} 7-7-7 \masalgo{} & \textbf{10.5} & \textbf{4.3} \\

\midrule
\texttt{Qwen-2.5-Coder-14B} & &  \\
\base{} 14 & 4.3 & 2.2 \\
\base{} 14 \saft{} & \textbf{16.8} & \textbf{5.8} \\
\mas{} 14-14-14 & 10.1 & 3.6 \\
\mas{} 14-14-14 \masft{} & 13.7 & 5.0 \\
\mas{} 14-14-14 \masalgo{} & 14.1 & \textbf{5.8} \\

\midrule
\texttt{Llama-3.1-8B} & &  \\
\base{} 8 & 2.2 & 0.0 \\
\base{} 8 \saft{} & 8.9 & 2.2 \\
\mas{} 8-8-8 & 2.2 & 0.0 \\
\mas{} 8-8-8 \masft{} & 8.9 & \textbf{4.3} \\
\mas{} 8-8-8 \masalgo{} & \textbf{10.8} & 3.6 \\

\midrule
\texttt{Phi-4-14B} & &  \\
\base{} 14 & 6.2 & 2.2 \\
\base{} 14 \saft{} & 14.1 & 3.6 \\
\mas{} 14-14-14 & 12.2 & 4.3 \\
\mas{} 14-14-14 \masft{} & 12.5 & 7.2 \\
\mas{} 14-14-14 \masalgo{} & \textbf{17.8} & \textbf{8.6} \\


\bottomrule
\end{tabular}
\caption{Performance comparison in single-agent and multi-agent settings on \emph{Test Challenge} (Test-C) set for \texttt{Qwen-2.5-Coder-7B}, \texttt{Qwen-2.5-Coder-14B}, \texttt{Llama-3.1-8B}, and \texttt{Phi-4-14B}.}
\label{tab:test_C_results_7b_8b}
\vspace{-1em}
\end{table}

\begin{figure*}
    \centering
    \includegraphics[width=\linewidth]{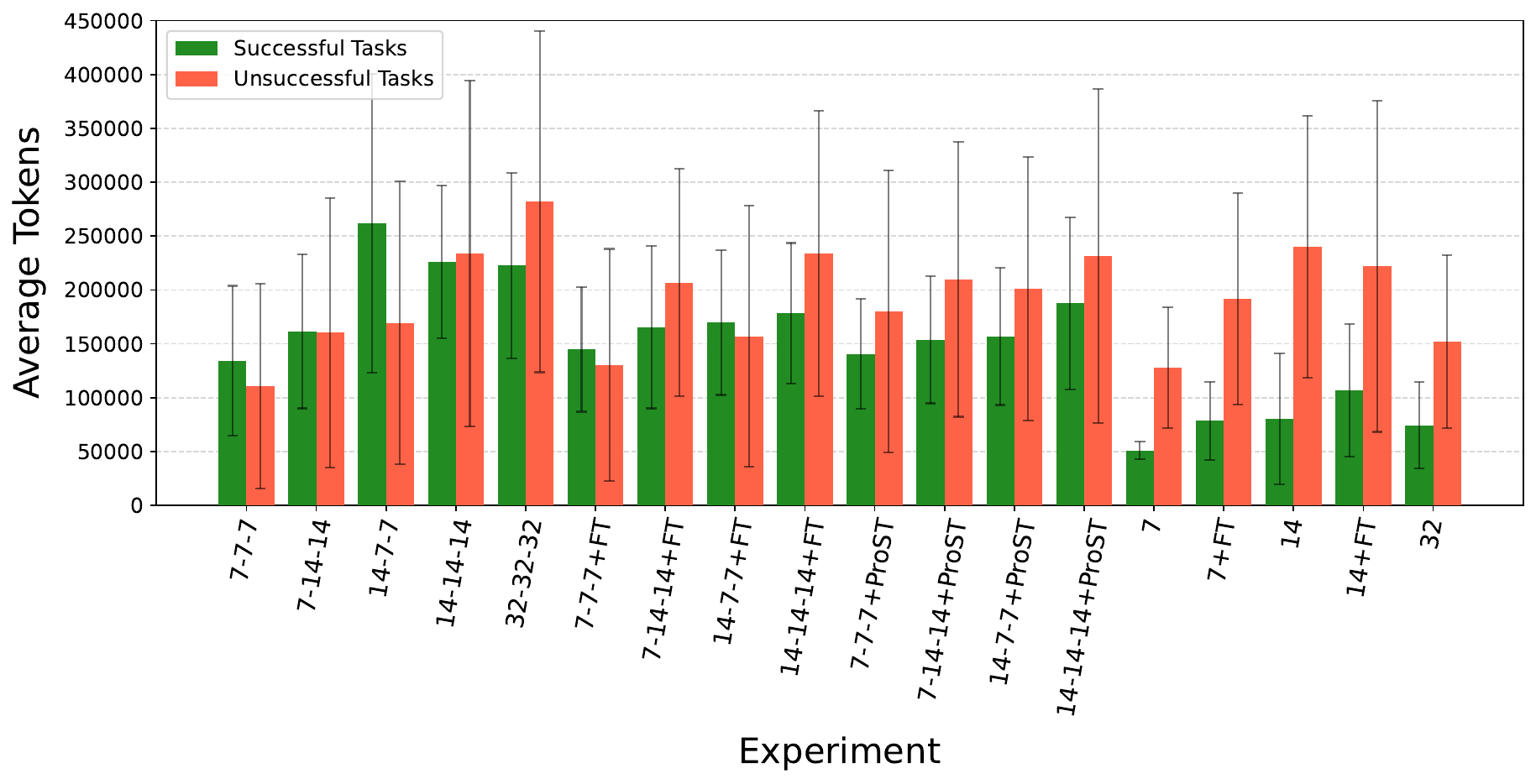}
    \caption{Comparison of average tokens across successful and unsuccessful tasks for all experiments. Green bars indicate average tokens spent on successful tasks, and red bars represent average tokens spend on each failed task.}
    \label{fig:tokens_count_all}
\end{figure*}

\section{Extended results}

\subsection{Generalizability of \algo{}}
\label{appendix:prost_generalizability}
The core components of our approach—role specialization, progressive subtask exposure, and Pareto-aware optimization—are not tied to AppWorld’s specific applications. The Orchestrator-Executor-Critic architecture is a generalizable decomposition pattern. Our training strategy, \algo{}, targets a universal challenge: SLMs’ inability to learn long trajectories under standard fine-tuning. Our ablation studies confirm that the order of subtask introduction matters, suggesting that curriculum design is a critical lever—not just an artifact of AppWorld’s tasks. Furthermore, we successfully apply \algo{} to non-coding models (Llama-3.1-8B, Phi-4-14B), showing consistent gains across architectures, and observe performance improvements on out-of-distribution tasks, where even modest gains indicate robustness. 
Using a single, high-quality benchmark for comprehensive evaluation is standard practice for complex problems due to a range of reasons. Firstly, working on complex task domains often requires significant engineering effort and environment setup cost. Many prominent works in software engineering and agent evaluation use single long horizon domains for this reason. For instance, the widely-cited SWE-bench \cite{jimenez2024swebench} paper has become the gold standard for evaluating coding agents and notable works~\citep{yang2024sweagentagentcomputerinterfacesenable, zhang2024autocoderoverautonomousprogramimprovement, ma2024lingmaswegptopendevelopmentprocesscentric, pan2025trainingsoftwareengineeringagents, golubev2024search, xia2024agentlessdemystifyingllmbasedsoftware, xie2025swefixertrainingopensourcellms} attempt to resolve programming tasks by interactive coding agents. Making progress on AppWorld represents useful advances in the complex problem space. The complex tasks in AppWorld are defined over every-day apps in the AppWorld benchmark, which is comparable to SWE-bench in scope and difficulty (see AppWorld details in \autoref{appendix:appworld_details}). This is a challenging benchmark and there are other recent works that focus exclusively on improving performance on this benchmark as well \citep{chen2025reinforcementlearninglonghorizoninteractive, gupta2025leveraging}. A recent multi-agent analysis paper also showed that AppWorld tasks pose substantial reasoning and coding challenges \cite{cemri2025multiagentllmsystemsfail}. Another recent work \cite{qi2025webrltrainingllmweb} evaluate solely their method on a long-horizon Web benchmark WebArena due to the challenges of complexity of configuring a second long horizon benchmark. Another limiting factor in exploring other complex tasks benchmarks is scacity of training data for multi agent systems. We spent a significant amount of time creating multi step reasoning training data (i.e for AppWorld) following rigorous processes. Thus, we argue that applying any other long-horizon benchmark’s tasks to the \algo{} training paradigm would also yield significant performance improvements compared to standard fine-tuning. 




\section{Prompts Details}

\label{appendix:prompts}

\begin{figure}[!htbp]
\begin{tcolorbox}
\tiny
\textbf{When creating subtasks, follow these guidelines:}
\begin{itemize}
    \setlength\itemsep{-1mm}
    \item Provide a clear description of the subtask, including the goal. Add every small detail from the main actual task.
    \item List the steps in natural language format to achieve the subtask's goal.
    \item Specify the apps that can be used to achieve the subtask's goal. \textbf{Do not include any API names}, as the orchestrator does not have knowledge of API names.
    \item Instruct the Executor agent to find relevant APIs from the API list of the specified app.
    \item Suggest checking the detailed documentation of the relevant API using \texttt{apis.api\_docs.show\_api\_doc(app\_name, api\_name)} to understand its arguments and output structure.
    \item Suggest calling the API using \texttt{apis.app\_name.api\_name} with the required parameters to get the desired information or perform the action. Mention detailed substeps (how to implement the logic) in natural language to achieve the subtask's goal.
    \item If the API requires authentication, suggest using the app's access token obtained from the previous authentication step.
\end{itemize}
\end{tcolorbox}
\caption{Prompt for the \orchestrator{} agent. Here we show only the extended part. We customized the original Re-Act prompt shown in Figure~\ref{prompt:single_agent_traj_generation}.}
\label{prompt:orchestrator}
\end{figure}

\begin{figure}[!htbp]
\begin{tcolorbox}
\tiny
\textbf{Key instructions:}
\begin{itemize}
    \setlength\itemsep{-1mm}
    \item For each step you take to solve the subtask, first write your thought process, then include code inside a \texttt{<code> ... </code>} block.
    \item Once you have completed the given subtask, include a summary of what you accomplished in your response.
    \begin{itemize}
        \item Mention all the variables’ exact names you used to store information and describe what they contain.
        \item Must end with the exit command in a \texttt{<code>exit</code>} block to signal completion. If you cannot solve the subtask, include a message explaining why and add a \texttt{<code>exit</code>} block. Do not use any other method to signal completion of the subtask.
    \end{itemize}
\end{itemize}
\end{tcolorbox}
\caption{Prompt for the \exec{} agent. Here we show only the extended part. We customized the original Re-Act prompt shown in Figure~\ref{prompt:single_agent_traj_generation}.}
\label{prompt:executor}
\end{figure}

\begin{figure}[!htbp]
\begin{tcolorbox}
\tiny
\textbf{Key instructions:}
\begin{itemize}
    \setlength\itemsep{-1mm}
    \item \textbf{REVIEW SCOPE:} Analyze the most recent step's code and reasoning and previous steps in the conversation. Look for:
    \begin{itemize}
        \item Logical flaws that would cause incorrect results with respect to the subtask.
        \item Whether the code snippet uses any undefined variables. If the variable is defined in previous steps, then we can use it.
        \item Whether the APIs are using correct parameter(s) according to the API documentation.
        \item Whether the code uses valid email addresses, access tokens, and variables from the actual conversation context, not placeholders.
    \end{itemize}   
    \item When reviewing code, check if variables from previous steps are being used correctly. Variables defined in earlier code blocks should be available in subsequent blocks.
    \item Ensure the code properly utilizes the \texttt{"supervisor"} app for account information and the \texttt{"phone"} app for contacts/friends/family data when needed.
    \item Check that the code examines API specifications using \texttt{apis.api\_docs.show\_api\_doc} before making API calls, and verify the API calls match the documented parameters exactly.
    \item Do not suggest code enhancements or optimizations. Focus solely on correctness and adherence to the subtask requirements.
    \item For APIs that return paginated results, verify the code properly loops through all pages using \texttt{page\_index} or a similar mechanism.
    \item Your role is to review code and provide feedback, not to write code. Focus on identifying issues as defined in \textbf{REVIEW SCOPE} and provide brief actionable feedback in natural language.     
\end{itemize}
\end{tcolorbox}
\caption{Prompt for the \critic{} agent. Here we show only the extended part. We customized the original Re-Act prompt shown in Figure~\ref{prompt:single_agent_traj_generation}.}
\label{prompt:critic}
\end{figure}

\begin{figure*}[!t]
\begin{tcolorbox}
\tiny
\textbf{USER:}\\
I am your supervisor and you are a super intelligent AI Assistant whose job is to achieve my day-to-day tasks completely autonomously. To do this, you will need to interact with app/s (e.g., spotify, venmo, etc) using their associated APIs on my behalf. For this you will undertake a *multi-step conversation* using a python REPL environment. That is, you will write the python code and the environment will execute it and show you the result, based on which, you will write python code for the next step and so on, until you've achieved the goal. This environment will let you interact with app/s using their associated APIs on my behalf.\\

To explore available APIs and functionality, following are the key commands:\\

To get a list of available apps\\
print(apis.api\_docs.show\_app\_descriptions())\\

To get the the list of available APIs in a specific app, e.g. supervisor\\
print(apis.api\_docs.show\_api\_descriptions(app\_name=`supervisor'))\\

To get the specification of a particular api, e.g. supervisor app's show\_account\_passwords\\
print(apis.api\_docs.show\_api\_doc(app\_name='supervisor', api\_name='show\_account\_passwords'))\\

To call a particular API from an app\\
print(apis.app\_name.api\_name(args))\\

To call the task completion API\\
print(apis.supervisor.complete\_task(answer=<answer>))\\

Each code execution will produce an output that you can use in subsequent calls. Using these APIs, you can now generate code, that I will execute, to solve the task. Let's start with the task\\

[Re-Act style example trajectory placeholder] \\

\textbf{USER:}\\
Congratulations, we solved the example task successfully. Now, before going to the next task, I want you to know the key instructions about solving the subtasks of the next task.\\

\textbf{Key instructions:}\\
\vspace*{-4mm}
\begin{itemize}
    \setlength\itemsep{-3mm}
    \item Remember that the email addresses, access tokens and variables (e.g. spotify\_password, spotify\_access\_token) in the example above are not valid anymore.\\
    \item For each step you take to solve the subtask, first write your thought process, then include code inside a <code> ... </code> block.\\
    \item Only generate valid code blocks, i.e., do not put them in ```...``` or add any extra formatting. \\
    \item Remember you can use the variables in your code in subsequent code blocks.\\
    \item Write small chunks of code and only one chunk of code in every step. Make sure everything is working correctly before making any irreversible change.\\
    \item The provided Python environment has access to its standard library. But modules and functions that have a risk of affecting the underlying OS, file system or process are disabled. You will get an error if do call them.\\
    \item Any reference to a file system in the task instructions means the file system *app*, operable via given APIs, and not the actual file system the code is running on. So do not write code making calls to os-level modules and functions.\\
    \item To interact with apps, only use the provided APIs, and not the corresponding Python packages. E.g., do NOT use `spotipy` for Spotify. Remember, the environment only has the standard library.\\
    \item The provided API documentation has both the input arguments and the output JSON schemas. All calls to APIs and parsing its outputs must be as per this documentation.\\
    \item Many APIs return items in "pages". Make sure to run through all the pages by looping over `page\_index`. For that use while true loop and check if the API returns any items. If it does, process them and increment the `page\_index` by 1. If it does not return any items, break the loop.\\
    \item If no direct API exists for the information you are looking for, examine all available APIs (relevent APIs first).\\
    \item Maintain variable names that are meaningful and relevant to the context of the task. Avoid using generic names like `result`, `data`, or `temp`. Instead, use descriptive names that reflect the content or purpose of the variable, such as `song\_list`, `spotify\_login\_result`, `song\_details`, `gmail\_password`, etc.\\
    \item To obtain current data or time, use Python functions like `datetime.now()` or obtain it from the phone app. Do not rely on your existing knowledge of what the current date or time is.\\
    \item For all temporal requests, use proper time boundaries, e.g., if I ask for something that happened yesterday, make sure to consider the time between 00:00:00 and 23:59:59. All requests are concerning a single, default (no) time zone.\\
    \item Any reference to my friends, family or any other person or relation refers to the people in my phone's contacts list.\\
    \item All my personal information, and information about my app account credentials, physical addresses and owned payment cards are stored in the "supervisor" app. You can access them via the APIs provided by the supervisor app.\\
    \item Once you have completed the task, call `apis.supervisor.complete\_task()`. If the task asks for some information, return it as the answer argument, i.e. call `apis.supervisor.complete\_task(answer=<answer>)`. For tasks that do not require an answer, just skip the answer argument or pass it as None.\\
    \item The answers, when given, should be just entity or number, not full sentences, e.g., `answer=10` for "How many songs are in the Spotify queue?". When an answer is a number, it should be in numbers, not in words, e.g., "10" and not "ten".\\
    \item You can also pass `status="fail"` in the complete\_task API if you are sure you cannot solve it and want to exit.\\
    \item You must make all decisions completely autonomously and not ask for any clarifications or confirmations from me or anyone else.\\
\end{itemize}

\textbf{USER:}\\
Using these APIs, now generate code to solve the actual task:\\

Supervisor name is: <first\_name> <last\_name>. Email is <email> and phone number is <phone\_number>.\\

Task: <task\_instruction>

\end{tcolorbox}
\caption{Prompt for single Re-Act agent inference}
\label{prompt:single_agent_traj_generation}
\end{figure*}

\begin{figure*}[t]
\centering
\begin{tcolorbox}[width=\textwidth]
\tiny
\begin{multicols}{2}
I am your supervisor and you are a highly intelligent AI Assistant. Your task is to transform a single-agent trajectory into a multi-agent trajectory for tasks in the AppWorld environment. The single-agent trajectory sucessfully resolved the task, so you need to ensure that the multi-agent trajectory also resolves the task successfully if I run the steps by order in the environment.\\
\newline
\# AppWorld Environment Overview\\
AppWorld is a simulated environment with 9 day-to-day apps that mimic real-world applications. This environment provides these following apps through the `apis` object:
\vspace*{-2mm}
\begin{itemize}
    \vspace*{-4mm}
    \setlength\itemsep{-1.5mm}
    \item amazon: Shopping and order management
    \item spotify: Music streaming and playlist management
    \item gmail: Email communication and management
    \item todoist: Task and to-do list management
    \item simple\_note: Note-taking and organization
    \item venmo: Person-to-person payments
    \item splitwise: Expense tracking and settlement
    \item file\_system: File operations and management
    \item phone: Calling and messaging functionality
\end{itemize}
\vspace*{-2mm}
Additionally, there are 2 helper applications:\\
\vspace*{-4mm}
\begin{itemize}
    \setlength\itemsep{-1.5mm}
    \item api\_docs: Provides interactive documentation lookup for all apps
    \item supervisor: Provides access to personal information (addresses, payment cards, account passwords)
\end{itemize}
\vspace*{-2mm}
\# Key API Commands\\
\vspace*{-4mm}
\begin{itemize}
    \setlength\itemsep{-1.5mm}
    \item `apis.api\_docs.show\_app\_descriptions()`: List available apps
    \item `apis.api\_docs.show\_api\_descriptions(app\_name=<app\_name>)`: List APIs for an app
    \item `apis.api\_docs.show\_api\_doc(app\_name=<app\_name>, api\_name=<api\_name>)`: Get API details
    \item `apis.app\_name.api\_name(args)`: Call an API
    \item `apis.supervisor.complete\_task(answer=<answer>)`: Complete task
\end{itemize}
\vspace*{-2mm}
\# Multi-Agent Framework\\
Your transformation will involve two agents:\\
\vspace*{1mm}
1. **Orchestrator Agent**: \\
\begin{itemize}
    \vspace*{-5mm}
    \setlength\itemsep{-1.5mm}
    \item Reviews the task and determines the next logical subtask
    \item Provides detailed subtask descriptions to the Executor
    \item Receives completion reports from the Executor
\end{itemize}
\vspace*{-2mm}
2. **Executor Agent**: \\
\vspace*{-2mm}
\begin{itemize}
    \vspace*{-3mm}
    \setlength\itemsep{-1.5mm}
    \item Performs the subtasks defined by the Orchestrator
    \item Not aware of the overall task, only focused on the current subtask
    \item Works in a Python REPL environment executing code step-by-step to accomplish the subtask
    \item Reports back to the Orchestrator upon subtask completion
\end{itemize}
\# Transformation Guidelines \\
\vspace{-1mm}
\#\# Subtask Design\\
\vspace*{-4mm}
\begin{itemize}
    \setlength\itemsep{-1.5mm}
    \item Create meaningful, logical subtasks that progress toward the overall goal
    \item Each subtask should be a discrete step toward completing the task
    \item Supervisor is the user, so use 'user' when talking about the who is using the app, for example "Find the song user liked" rather than "Find the song 'I', or 'you' liked"
    \item Authentication/login to an app should always be a separate subtask
    \item Must ensure the final subtask involves calling the task completion API. Orchestrator agent should instruct to call `apis.supervisor.complete\_task()`. If the task requires information, should ask to return it using the answer parameter: `apis.supervisor.complete\_task(answer=<answer>)`.
\end{itemize}
\#\# Subtask Description Format \\
\vspace*{-4mm}
\begin{itemize}
    \setlength\itemsep{-1.5mm}
    \item Begin with a brief description of the subtask's goal
    \item List all possible logical steps in details by order to accomplish the subtask. The order of the steps should be the same as the order of the steps in the original single-agent trajectory
    \item For authentication related subtasks, include these specific steps:
    \vspace*{-2mm}
    \begin{itemize}
        \setlength\itemsep{-0.5mm}
        \item Add supervisor name, email, and phone number to the description as this info will be helpful for the Executor agent for authentication
        \item Suggest to explore the API documentation of the app using `apis.api\_docs.show\_api\_descriptions (app\_name=<app\_name>)` to find the authentication-related API
        \item Suggest to check the detailed documentation of the authentication API using `apis.api\_docs.show\_api\_doc (app\_name=<app\_name>, api\_name=<api\_name>)` to understand its arguments and output structure.

        \item For username, suggest to use username (e.g., email address, phone number, etc.) given in the subtask description
        \item For password, suggest to find the account's password retrieval API from the "supervisor" app and call the API to retrieve the password
        \item Suggest to call the login API using `apis.<app\_name>.<login\_api> (username=<username>, password=<password>)` with the collected username and password. And store the token for future use
    \end{itemize}
    
    \item For other subtasks,
    \vspace*{-2mm}
    \begin{itemize}
        \setlength\itemsep{-.5mm}
        \item List the steps in natural language format
\columnbreak

        \item Specify the possible apps that can be used to achieve the subtask's goal. Don't include any API names as orchestrator don't have knowledge of the API names
        \item **Don't include actual API names in the description, instead suggest to explore the API documentation using `apis.api\_docs.show\_api\_descriptions (app\_name=<app\_name>)` to find the relevant API. Then suggest to findout relevent APIs**
        \item Suggest to check the detailed documentation of the relevent API using `apis.api\_docs.show\_api\_doc (app\_name=<app\_name>, api\_name=<api\_name>)` to understand its arguments and output structure.
    \end{itemize}
    \item In the subtask description, include actual path, file or people names, if mentioned in the task description. For example, mention actual path instead of saying "...the specified path..."
    \item For final subtask, **must include a note to call the task completion API using `apis.supervisor.complete\_task(answer=<answer>)` to signal that the overall task has been completed. If the task requires information, should ask to return it using the answer parameter: `apis.supervisor.complete\_task(answer=<answer>)`**
    \item End with instruction to report back upon completion
\end{itemize}
\#\# Executor Steps Format\\
\vspace*{-4mm}
\begin{itemize}
    \setlength\itemsep{-1.5mm}
    \item Each step (thought and code) must be identical to the original step in the single-agent trajectory
    \item Must include both the thought and the code exactly as in the original step
    \item Must enclosed on code in <code> </code> tags
    \item Upon subtask completion, add a summary and exit command: <code>exit</code>. Summary should include what executor agent has accomplished in the subtask and signal to the Orchestrator agent that the subtask is complete.
\end{itemize}

\#\# Output Requirements\\
\vspace*{-4mm}
\begin{itemize}
    \setlength\itemsep{-1.5mm}
    \item IMPORTANT: You must respond ONLY with valid JSON. Do not include any explanatory text, introductions, or markdown formatting outside of the JSON object. Your entire response must be parseable as JSON.
    \item Follow the exact JSON structure shown in the below example. The example provided below is for illustrative purposes only. In the given example- the App, and API names were only for demonstration.
    \item Do not exclude any original step or code
    \item Do not invent new code except for the final exit command
    \item Preserve all original steps; it's thought as well as the code
    \item Divide into logical subtasks with authentication always as a separate subtask
    \item Define thought first and then format only code within <code> tags and end each subtask with <code>exit</code>
    \item Do not include actual API names in the description, instead suggest to explore the API documentation using `apis.api\_docs.show\_api\_descriptions(app\_name=<app\_name>)` to find the relevant API. Then suggest to findout relevent APIs. Then check the detailed documentation of the relevent API using `apis.api\_docs.show\_api\_doc(app\_name=<app\_name>, api\_name=<api\_name>)` to understand its arguments and output structure.
    \item Use the placeholder `<app\_name>` and `<api\_name>` in the subtask description, because the orchestrator agent doesn't know the API names
    \item In final subtask, include a note to call the task completion API using `apis.supervisor.complete\_task(answer=<answer>)` to signal that the overall task has been completed. If the task requires information, should ask to return it using the answer parameter: `apis.supervisor.complete\_task(answer=<answer>)`
\end{itemize}

Your goal is to transform a single-agent trajectory into a multi-agent trajectory following the JSON schema below:\\
\vspace*{-2mm}
\begin{verbatim}
{
  "subtasks": [
    {
      "subtask_number": <integer>,
      "subtask_description": <string>,
      "executor_steps": [
        {
          "subtask_number": <integer>,
          "step_number": <integer>,
          "plan_and_code": <string and <code> code </code> >
        },
      ]
    },
  ]
}
\end{verbatim}

For example, consider the following task and the multi-agent trajectory solution where two agents are solving the task.\\

[Example Multi-Agent Trajectory Placeholder]\\

Now translate the actual single-agent trajectory into a multi-agent trajectory:\\

Actual task: <task\_description>\\
Supervisor name is: <first\_name> <last\_name>. Email is <email> and phone number is <phone\_number>.\\

The single-agent trajectory is as follows: <single\_agent\_trajectory>

\end{multicols}
\end{tcolorbox}
\caption{Prompt for Single-agent to multi-agent trajectory transformation}
\label{fig:sat_to_mat}
\end{figure*}

\end{document}